\documentclass{article}
\usepackage{amssymb}

\usepackage[final]{corl_2025} 

\usepackage{graphics} 
\usepackage{amsmath} 
\usepackage{amssymb}  
\usepackage{caption}
\usepackage{hyperref}
\usepackage{algorithm}
\usepackage{algorithmic}
\usepackage{multirow}
\usepackage{booktabs}
\usepackage{pifont}
\usepackage{enumitem}
\usepackage{tcolorbox}
\usepackage{makecell}
\usepackage{multicol}
\usepackage{wrapfig}
\usepackage[title]{appendix}

\newcommand{\cmark}{\ding{51}}%
\newcommand{\xmark}{\ding{55}}%

\def\eg{\textit{e.g}~}

\def\ie{\textit{i.e}~}

\title{exUMI: Extensible Robot Teaching System with Action-aware Task-agnostic Tactile Representation}

\author{
  Yue Xu$^1$,~Litao Wei$^1$,~Pengyu An$^1$,~Qingyu Zhang$^1$,~Yong-Lu Li$^{1,2}$\thanks{Corresponding author.} \\
  $^1$Shanghai Jiao Tong University, $^2$Shanghai Innovation Institute \\
  \texttt{\{silicxuyue,oscar0731,anpengyu,leozhangchina,yonglu\_li\}@sjtu.edu.cn} \\
}

\begin{document}
\maketitle


\begin{abstract}
    Tactile-aware robot learning faces critical challenges in data collection and representation due to data scarcity and sparsity, and the absence of force feedback in existing systems. To address these limitations, we introduce a tactile robot learning system with both hardware and algorithm innovations. We present \textbf{exUMI}, an extensible data collection device that enhances the vanilla UMI with robust proprioception (via AR MoCap and rotary encoder), modular visuo-tactile sensing, and automated calibration, achieving 100\% data usability. Building on an efficient collection of over \textbf{1 M} tactile frames, we propose Tactile Prediction Pretraining (\textbf{TPP}), a representation learning framework through action-aware temporal tactile prediction, capturing contact dynamics and mitigating tactile sparsity. Real-world experiments show that TPP outperforms traditional tactile imitation learning. Our work bridges the gap between human tactile intuition and robot learning through co-designed hardware and algorithms, offering open-source resources to advance contact-rich manipulation research. \\
    \textbf{Project page}: \url{https://silicx.github.io/exUMI}.
\end{abstract}



\begin{figure}[h]
    \includegraphics[width=\linewidth]{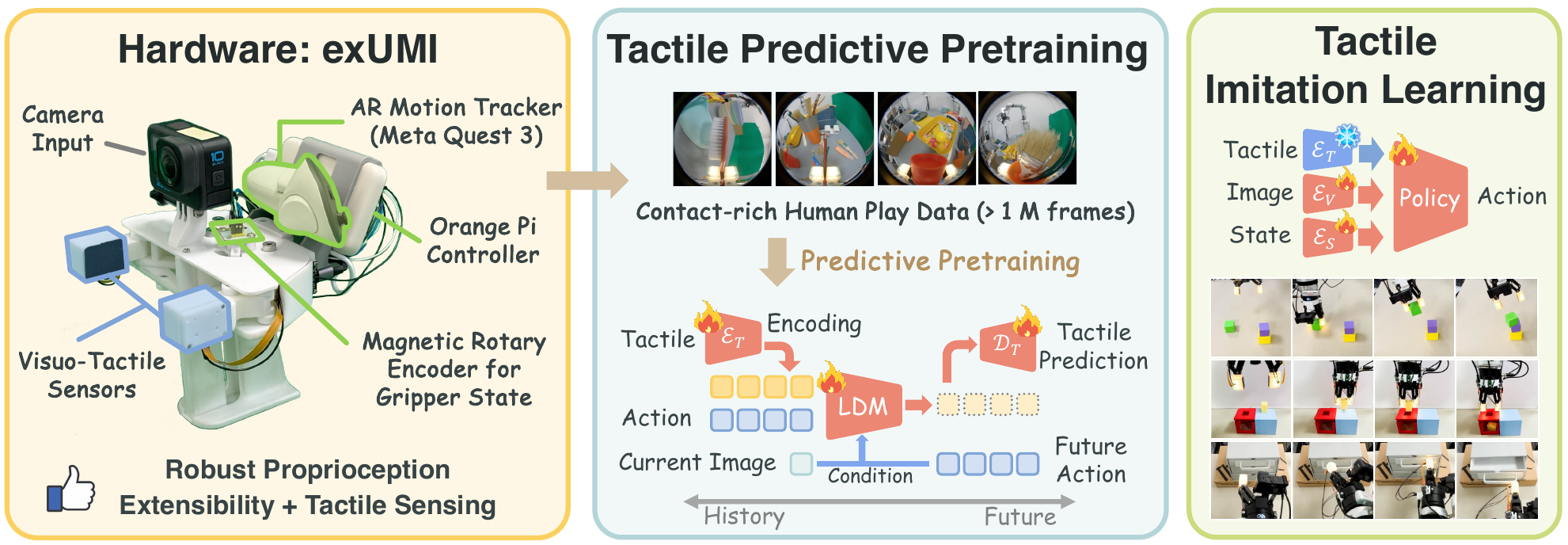}
    \captionof{figure}{
    We present a co-design of hardware and algorithm for tactile-aware robot learning. We present \textbf{exUMI}, an extensible upgrade for the UMI~\cite{umi} system (left). With the hardware, we learn tactile representation by temporal tactile prediction (middle), encoding the tactile dynamics conditioned on robot action. We evaluate our representation on multiple real-world robotic tasks (right).
    }
    \vspace{-6px}
    \label{fig:teaser}
\end{figure}

\section{Introduction}

Collecting robot manipulation data is crucial for developing autonomous robots in real-world environments. While teleoperation techniques~\cite{zhu2023viola,mandlekar2018roboturk,luo2024human} yield accurate robot learning data, they are labor-intensive, inefficient, and expensive. Learning from human demonstration~\cite{shao2021concept2robot,chen2021learning,bahl2022human,simeonov2022neural,schmeckpeper2020learning,qin2022dexmv,xiao2022masked} is cheap and massive, but inaccurate for robot embodiment. 
Between these two extremes, portable hand-held devices offer a promising middle ground.
Recently, UMI~\cite{umi} was proposed to alleviate the deployment gap, which is a portable physical twin of the robot gripper for in-the-wild demonstration collection. It shows remarkable capabilities in collecting kinesthetic teaching data, enabling efficient transfer of human demonstrations to robotic systems.

However, when it comes to tactile-aware robot learning, traditional data collection systems face challenges. 
First, human demonstrators rely on tactile feedback to adjust manipulation strategies, making it hard or impossible for the teleoperation systems to collect demonstrations for force-sensitive tasks. 
Second, tactile signals in robot learning are severely sparse, with valid contacts occupying less than 10\% of manipulation trajectories~\cite {fu2024touch}. 
This undermines conventional tactile learning approaches: 
(1) Direct imitation learning~\cite{10611113,Lee_2024} suffers from data scarcity. 
(2) Self-supervised pretraining~\cite{rodriguez2024contrastivetouchtotouchpretraining,guzey2023dexteritytouchselfsupervisedpretraining,yu2025mimictouchleveragingmultimodalhuman,wu2025canonicalrepresentationforcebasedpretraining} learns with the potentially incorrect inductive bias (\eg, translation invariance). 
(3) Visual-tactile alignment paradigm~\cite{10802778,10610567,pattabiraman2024learningprecisecontactrichmanipulation} overlooks that vision and tactile modalities have a one-to-many relation considering the contact force. 
Therefore, the resulting representations often fail to generalize beyond narrow task-specific scenarios.

To address these challenges, we present a solution of tactile representation learning with the co-design of both novel hardware and an algorithm.
We propose \underline{\bf exUMI} (Sec.~\ref{sec:hardware}), an \underline{\bf ex}tensible upgrade to the \underline{\bf UMI} robot teaching system~\cite{umi}.
exUMI is a portable hand-held data collection device with visuo-tactile sensors and a motion capture system for teaching robot trajectory.
Our system has three key innovations: 
(1) A robust proprioception subsystem of AR-based motion capture (Meta Quest 3) and magnetic rotary encoder (AS5600), achieving nearly 100\% data usability by replacing the vulnerable SLAM and ArUco systems.
(2) A central controller enabling modality extensibility, coupled with temporal sensor alignment protocols during human demonstrations with $<$50~ms error. 
(3) Visuo-tactile integration of an upgraded design of 9DTact~\cite{9dtact} for stable quality control and better durability. 
With exUMI, users could efficiently collect robot learning and tactile sensing data with the least effort. For a simple pick-and-place task, a user could collect \textbf{100} demonstrations in \textbf{20} minutes to achieve \textbf{100\%} data usability and over \textbf{70\%} task success rate by behavior cloning.

Building on the hardware basis, we propose an \textit{action-aware} but \textit{task-agnostic} tactile representation learning framework (Sec.~\ref{sec:method}). 
To exploit exUMI's unique data properties and address the previously mentioned potential issues, we propose \textbf{T}actile \textbf{P}redictive \textbf{P}retraining (\textbf{TPP}), learning tactile representation by the proxy task of \textbf{temporal tactile prediction}. 
We pretrain a tactile diffusion model to predict future tactile frames, conditioned on the action sequence and the current camera image.
The representation is learned considering the physical action on the sensor, mirroring human haptic perception, where contact dynamics could be inferred from future movement. 
To enable the action-aware pretraining, we use exUMI to efficiently collect large-scale \textit{human play} data by randomly interacting with objects, producing a contact-rich tactile-action aligned dataset of $>$\textbf{1~M frames} with over \textbf{10 times} efficiency than teleoperation. 
The pretrained representation model could be embedded in an imitation learning policy, and our tactile embeddings empirically achieve significant success gain in force-sensitive tasks (\eg, ``pull drawer, peg in hole'') versus vision-only baselines.

This work establishes a new paradigm for tactile-aware robot learning to overcome fundamental bottlenecks in tactile learning. 
Our primary contributions include:
(1) \textbf{exUMI}, a tactile robot data system that enhances UMI with 100\% reliable proprioception and tactile sensing;
(2) \textbf{Tactile Prediction Pretraining (TPP)}, an action-aware and task-agnostic tactile representation learning method that exploits contact dynamics through forward tactile prediction;
(3) A large-scale tactile-action aligned robot dataset with over 1~M frames;
(4) Empirical validation showing over 20\% performance gains over tactile learning baselines.

\section{Related Work}
\label{sec:related}

\textbf{Robot Data Collection Systems} are essential for training generalizable manipulation policies. Traditional approaches include teleoperation~\cite{fan2023digital,luo2024human,chi2023diffusion,seo2023deep}, which offers high-quality data but is costly; and human data based methods~\cite{shao2021concept2robot,chen2021learning,bahl2022human,ma2022vip,bahl2023affordances,simeonov2022neural,pan2023tax,shen2023distilled,schmeckpeper2020learning,qin2022dexmv,nair2022r3m,xiao2022masked}, which lack precision. Portable systems like UMI~\cite{umi} and AirExo~\cite{fang2024airexo} employ a handheld or exoskeleton as a physical twin of a robot, enabling efficient in-the-wild demonstrations. Recent variants address existing limitations: Fast-UMI~\cite{wu2024fast} enhances the SLAM system of UMI with RealSense T265, while ForceMimic~\cite{liu2024forcemimic} integrates a force-torque sensor for contact-rich tasks. These advancements improve scalability, adaptability, and multimodal data acquisition for robotic learning.

\textbf{Tactile Representation Learning} aims to extract meaningful tactile features for robots to perceive geometry properties and interaction dynamics. Current trending involves:
(1) \textit{Directly imitation training}: minimally process the tactile data~\cite{10611113} or even directly use it without any processing~\cite{Lee_2024} for training via reinforcement learning methods. Further learning process involves modality fusion models such as transformer~\cite{10802778,10610567,pattabiraman2024learningprecisecontactrichmanipulation}.
(2) \textit{Intermediate Representation:} transform raw data into a manual representation space that captures task-relevant semantics, such as converting into point clouds~\cite{10610532,huang20253dvitaclearningfinegrainedmanipulation}, or reconstruct with NeRF~\cite{doi:10.1126/scirobotics.adl0628}.
(3) \textit{Self-Supervised Learning} (SSL): leverages the inherent structure and temporal-spatial correlations within raw tactile data through proxy tasks. \citet{rodriguez2024contrastivetouchtotouchpretraining} extends contrastive learning to paired tactile data. \citet{guzey2023dexteritytouchselfsupervisedpretraining,yu2025mimictouchleveragingmultimodalhuman} extend traditional SSL methods like BYOL to tactile tasks and train on play data or task-specific data. \citet{wu2025canonicalrepresentationforcebasedpretraining} utilizes the masked learning method.
\citet{feng2025anytouch} combine the pixel level SSL and contrastive learning to learn a cross-sensor tactile representation.
\citet{zhao2024transferable} exploit multimodal and multitask joint representation learning for a semantically meaningful tactile model.

\textbf{Real-World Tactile Datasets} have been developed to advance tactile perception and manipulation. Early progress in tactile sensing was driven by data collected using GelSight and, more recently, DIGIT sensors. Notable examples include Calandra \textit{et.~al.}~\cite{MoreThanFeel,FeelSucc}, SSVTP~\cite{kerr2022self}, datasets collected through a self-supervised automated process, respectively utilizing GelSight and DIGIT; VisGel~\cite{VisGel}, a synchronized vision-touch dataset comprising diverse everyday objects; X-Capture~\cite{clarke2025x}, adding depth information and acoustic information. Touch and Go~\cite{yang2022touchgolearninghumancollected}, a single model approach that utilizes a portable device, enabling human data collection and introducing a large diversity in scenes. 
In comparison, our dataset addresses the challenges of efficient collection and proprioception alignment, resulting in a significantly larger data volume than any existing dataset.


\begin{figure}[t]
    \includegraphics[width=\linewidth]{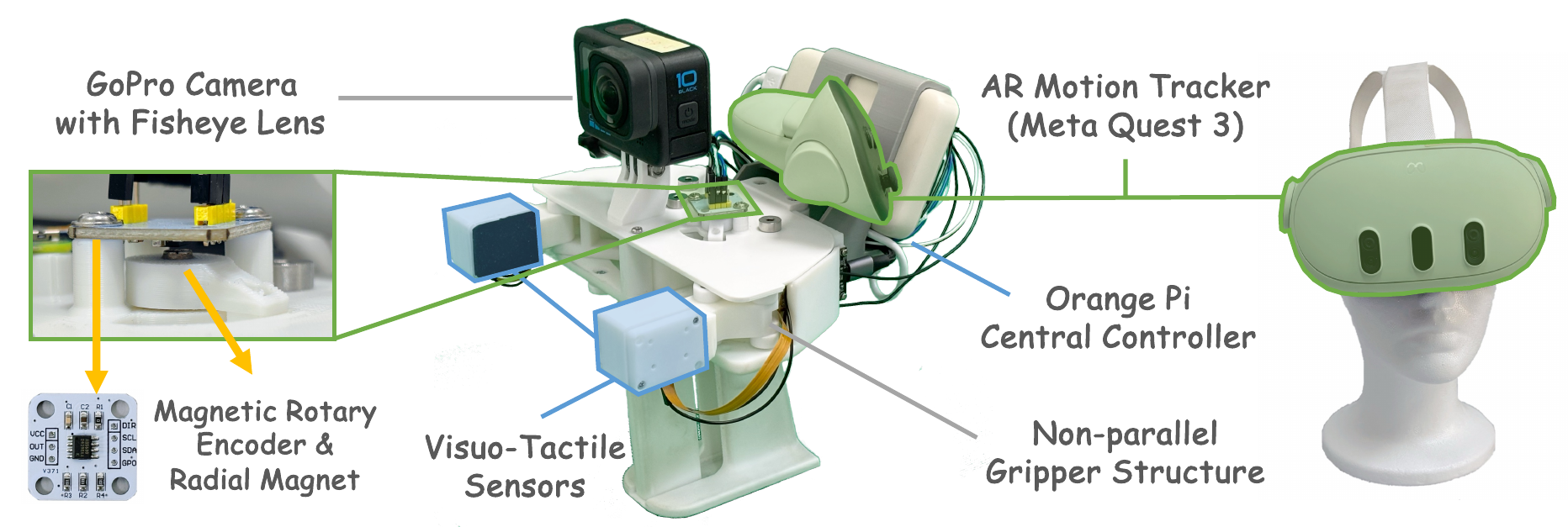}
    \captionof{figure}{
    \textbf{exUMI} hardware system. We extend the UMI framework by disentangling proprioception into an AR motion capture system and a rotary encoder for precise gripper width. 
    A central controller with automatic latency calibration enables additional sensor integration with maximal mobility.
    We attach two visuo-tactile sensors to the gripper fingertips.
    }
    \vspace{-5px}
    \label{fig:hardware-overview}
\end{figure}

\section{Hardware System}
\label{sec:hardware}

We present \textbf{exUMI}, an enhanced hardware design upon UMI~\cite{umi}, guided by three key principles:
\begin{itemize}[leftmargin=1em,itemsep=0pt,topsep=0pt]
\item \textbf{Precise robot proprioception}: Precise tracking of end-effector 6D pose and gripper width. The vanilla UMI system relies on visual SLAM and ArUco tracking, which is vulnerable (Fig.~\ref{fig:hard-case}).
\item \textbf{Extensibility}: Seamless integration of additional sensors through centralized control.
\item \textbf{Portability}: Ensuring in-the-wild data collection without fixed infrastructure (\eg, base station).
\end{itemize}
Therefore, we develop the hardware system with the following key components. Due to the page limit, please refer to Appendix~\ref{sec:hardware-detail} for full details.

\subsection{Hardware Design}

\textbf{Magnetic Rotary Encoder.}
The gripper state estimation in the original UMI design relies on ArUco markers, which suffer from occlusion and severe fisheye distortion, contributing to approximately 10\% of data preprocessing failures and notable errors.
To achieve accurate and robust gripper width tracking, we propose a low-cost AS5600 magnetic rotary encoder solution.
As shown in Fig.~\ref{fig:hardware-overview}, we modify the top plate to attach a radial magnet and the rotary sensor above a joint.

\begin{figure}[t]
    \centering
    \begin{minipage}[c]{0.31\linewidth}
        \centering
        \includegraphics[width=0.9\linewidth]{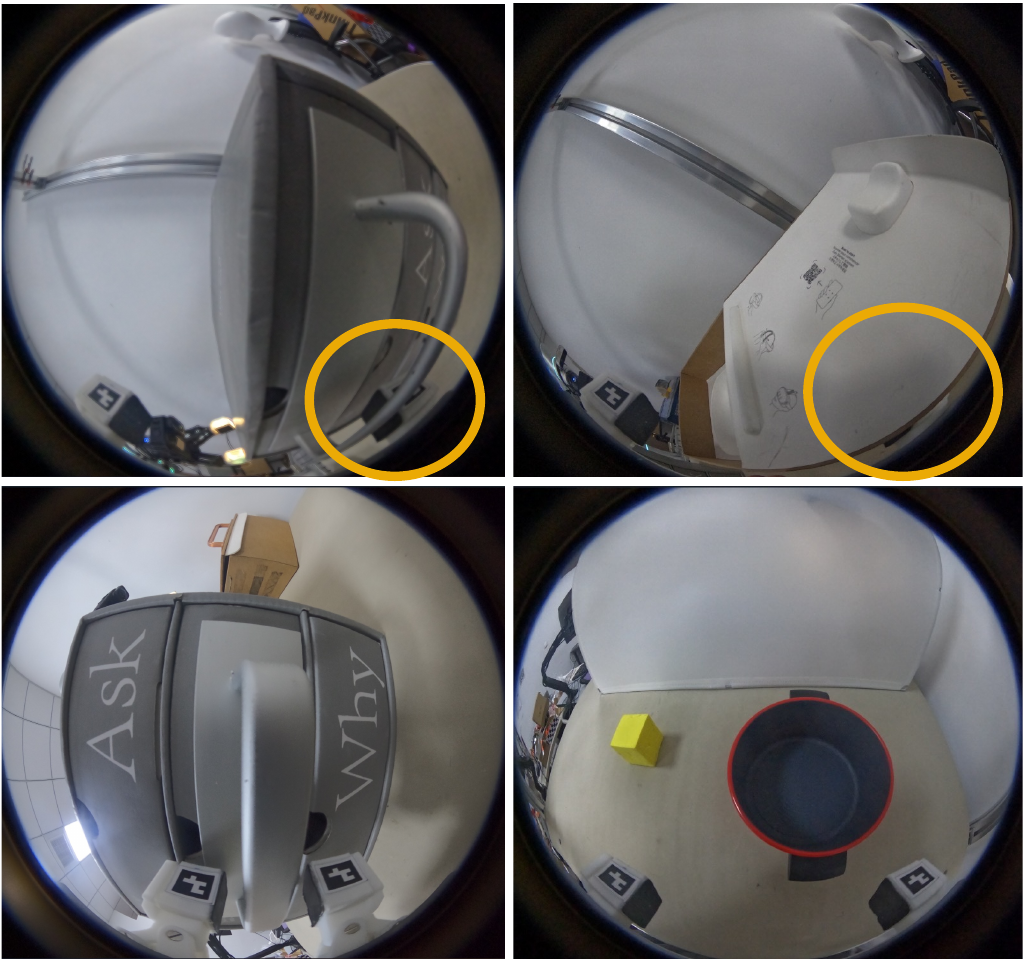}
        \caption{Hard scenarios for SLAM and marker detection (clean background, occlusion).}
        \label{fig:hard-case}
    \end{minipage}
    \hfill
    \begin{minipage}[c]{0.32\linewidth}
        \centering
        \includegraphics[width=0.8\linewidth]{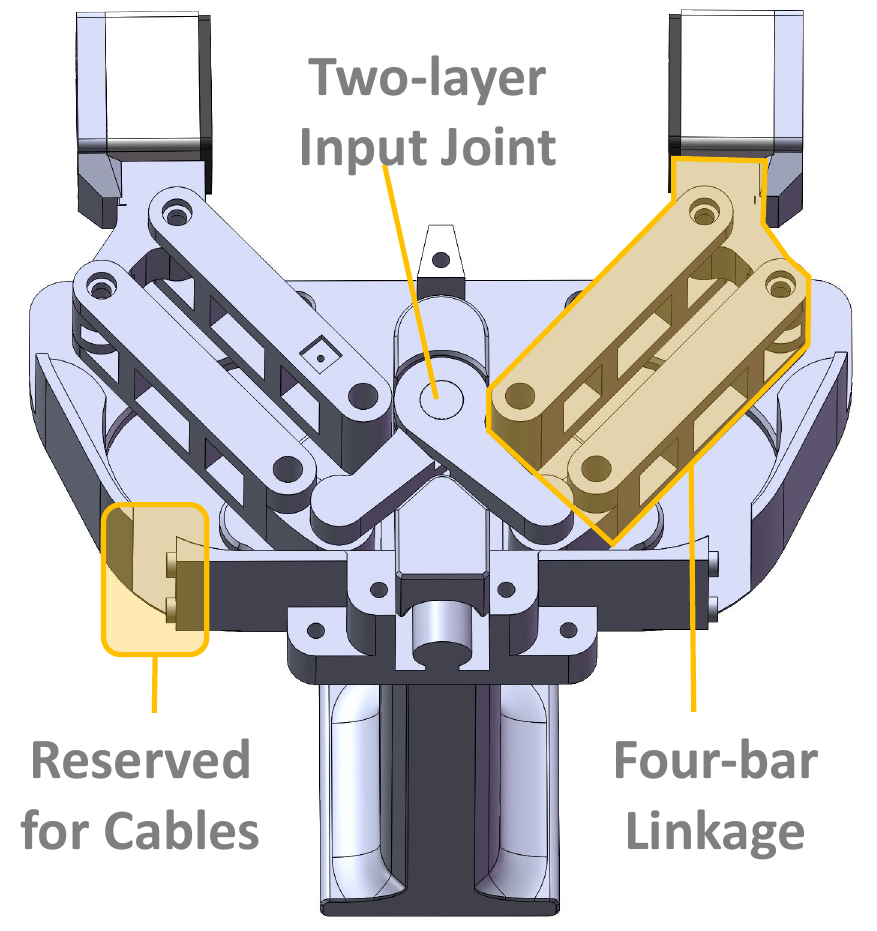}
        \caption{Non-parallel gripper mechanism.}
        \label{fig:mechanical}
    \end{minipage}
    \hfill
    \begin{minipage}[c]{0.34\linewidth}
        \centering
        \includegraphics[width=0.98\linewidth]{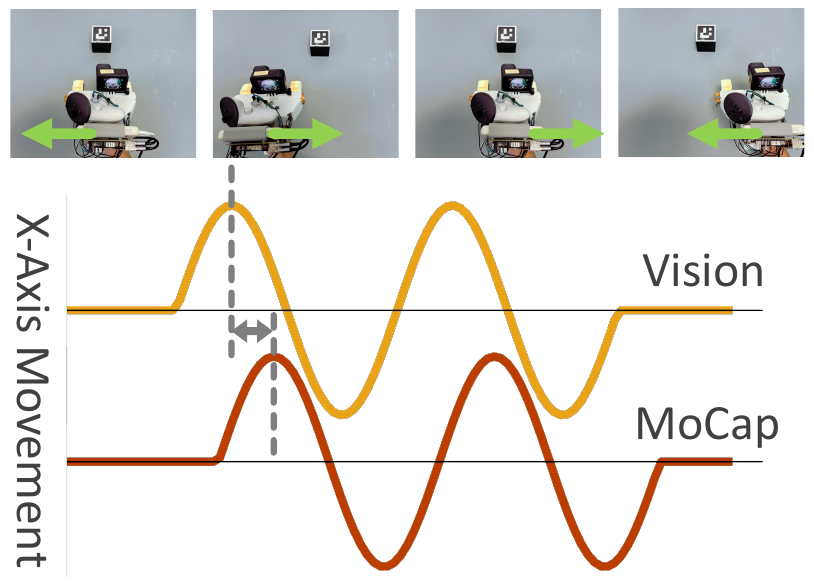}
        \caption{The latency calibration process of AR MoCap and RGB camera.}
        \label{fig:latency-align}
    \end{minipage}
    \vspace{-10px}
\end{figure}

\textbf{AR Motion Capture System.}
To overcome the limitations of vision-based tracking (SLAM) in occluded or complex scenarios, use a Meta Quest 3 headset for end-effector 6D pose capture following~\citet{chen2024arcap}, which is accurate and robust to occlusion.
We integrate the left VR controller through a custom-designed mount attached to the UMI body, which also provides space for the power supply and an Orange Pi controller, serving as a universal sensor hub, synchronously capturing data from the AR headset, rotary encoder, and any additional sensors, such as tactile sensors. 
Our MoCap system achieves less than 10 mm error on average compared to FastUMI~\cite{wu2024fast}.


\textbf{Fingertip Visuo-tactile Sensors.}
We embed visuo-tactile sensors on the fingertips for modality extension. We propose a low-cost visuo-tactile sensor based on 9DTact~\cite{9dtact}. We redesigned the sensor model with a contact protection structure, which could hold the silicon gel and secure it from large tangent forces. The cable and power connection are also optimized for durability. We also use a customized mold to ensure the same and stable thickness of the silicon gel for consistent tactile sensing.
The upgraded sensor design achieves significantly enhanced durability and stability.

\textbf{Visual Input.}
We employ the same GoPro setting to \citet{wu2024fast} for a wider and clearer view.

\textbf{Non-Parallel Gripper Mechanical Design.}
We design an additional mechanical system (Fig.~\ref{fig:mechanical}) for the users of non-parallel grippers such as Flexiv Grav and Robotiq 2F.

\textbf{Cost and Accessibility.}
Our system is low-cost and DIY-friendly with a default configuration starting at \textbf{\$ 698}, making it suitable for research/education. All CAD files will be released.

\subsection{Data Processing}

Our careful design enables robust in-the-wild data collection with minimal calibration overhead.

\textbf{System Calibration}.
Our system requires two \textbf{one-time} calibrations:
(1) {\it AR Controller Calibration}: the user aligns the exUMI with the base coordinates in the AR space and records the controller's transform, which is then used to correct the pose tracking.
(2) {\it Gripper State Calibration}: incrementally positioned the gripper at 1 cm intervals and recorded AS5600 reading, which is then interpolated and used for mapping from AS5600 readings to absolute gripper width.

\textbf{Latency Calibration.}
We designed a calibration protocol to synchronize AR motion capture (and the tactile signals) with the visual inputs (Fig.~\ref{fig:latency-align}). At the beginning of data collection, the user horizontally sweeps above an ArUco marker. We extract the x-axis movement of the AR MoCap and the marker trajectories, then use an MSE minimization algorithm to find the latency offset between the two curves. This approach ensures tight synchronization between the two systems.

Overall, exUMI could efficiently collect robot demonstrations with accurate 6D pose trajectory and gripper width, aligned RGB image input, and additional tactile signals. Our data collection pipeline is significantly simplified and more robust, leading to a nearly 100\% data processing success rate compared to less than 60\% of vanilla UMI. Our system could significantly enhance data efficiency and also enable multi-modality data collection at an affordable price.

\section{Methodology}
\label{sec:method}

\subsection{Taxonomy of Tactile Representation Learning}  
\label{sec:taxonomy}  

Tactile representation learning converts high-dimensional raw touch signals into compact features, which is fundamental to alleviating challenges like sensor heterogeneity, data dimension, data scarcity, and the need for real-world generalization.
Regularly, its target is to learn a tactile encoder $\mathcal{E}_T$ for further multimodal policy learning:
$\pi(\mathbf{a}_{t} | \mathcal{E}_S(\mathbf{s}_{t}), \mathcal{E}_T(\mathbf{T}_t), \mathcal{E}_V(\mathbf{V}_t))$,
where $\mathbf{a}$ is the action, $\mathbf{T}$, $\mathbf{V}$, $\mathbf{s}$ are tactile, visual and robot state inputs, and $\mathcal{E}_S$, $\mathcal{E}_T$, $\mathcal{E}_V$ are the encoders. 
Currently, we classify the learning methods of $\mathcal{E}_T$ to three paradigms, which will be detailed in Appendix~\ref{app:sec:taxonomy}.

\textbf{(a) Direct Multimodal Imitation Learning}~\cite{10611113,Lee_2024,10802778,10610567,pattabiraman2024learningprecisecontactrichmanipulation}: end-to-end learning $\mathcal{E}_T$ during the training of $\pi$, which suffers from tactile data scarcity since both the data size and the proportion of valid tactile contacts are limited.
\textbf{(b) Spatial Self-Supervised Learning}:
methods like contrastive learning~\cite{rodriguez2024contrastivetouchtotouchpretraining,guzey2023dexteritytouchselfsupervisedpretraining,yu2025mimictouchleveragingmultimodalhuman} and masked learning~\cite{wu2025canonicalrepresentationforcebasedpretraining} learn tactile embeddings $\mathcal{E}_T(\mathbf{T}_t)$ through proxy objectives. But these method usually imposes incorrect inductive biases borrowed from vision \eg, geometrical self-consistency and translation invariance, which may not exist in tactile sensing.
\textbf{(c) Visual-Tactile Alignment}~\cite{VisGel,clarke2025x}: learns joint embeddings by maximizing similarity $s\left(\mathcal{E}_T(\mathbf{T}_t), \mathcal{E}_V(\mathbf{V}_t)\right)$. It assumes a coarse \textit{one-to-one visuo-tactile mapping}, regardless of the actual \textit{one-to-many relation} when different contact forces are applied. 

Therefore, to overcome data scarcity of tactile sensing and for task transferability, representation pretraining (such as (b), (c)) is critical.
However, current pretraining approaches face limitations that stem from a shared oversight: treating tactile signals as static observations rather than \textit{action-aware dynamic processes}. Human tactile understanding intrinsically combines contact mechanics with motion intent (\eg, ``\textit{if I push harder, drag the object left, the slip risk decreases, and the tactile signal will be more significant}''). Our framework bridges this gap by reformulating tactile learning as an action-conditioned temporal prediction problem, explicitly modeling the forward tactile dynamics that underpin real-world contact interactions.

\subsection{Action-aware Tactile Data Collection}


For the tactile pretraining process, we efficiently collect tactile-action aligned human play data leveraging the portability of the exUMI system.
The collectors randomly manipulate diverse objects across \textbf{10} real-world environments, interacting with \textbf{300+} objects spanning from rigid tools to deformable fabrics and granular materials. 
Finally, we collect a total of \textbf{1 M} frames of aligned images-tactile-action data. Our data has rich contacts with over 60\% active tactile frames, compared to less than 10\% of regular data collection~\cite{fu2024touch}.
The contact richness further enhances our collection efficiency, and the \textbf{480 K} tactile frames are collected from just 5 hours of human interaction, which would take 10$\times$ the time for a teleoperation system.
Although having different purposes and granularity, the dataset is significantly larger than the previous tactile datasets (\eg, TVL~\citep{fu2024touch} has 43.7 K frames), which is sufficient for our tactile learning.

\subsection{Tactile Predictive Pretraining}

\begin{figure}[t]
\centering
\includegraphics[width=0.95\textwidth]{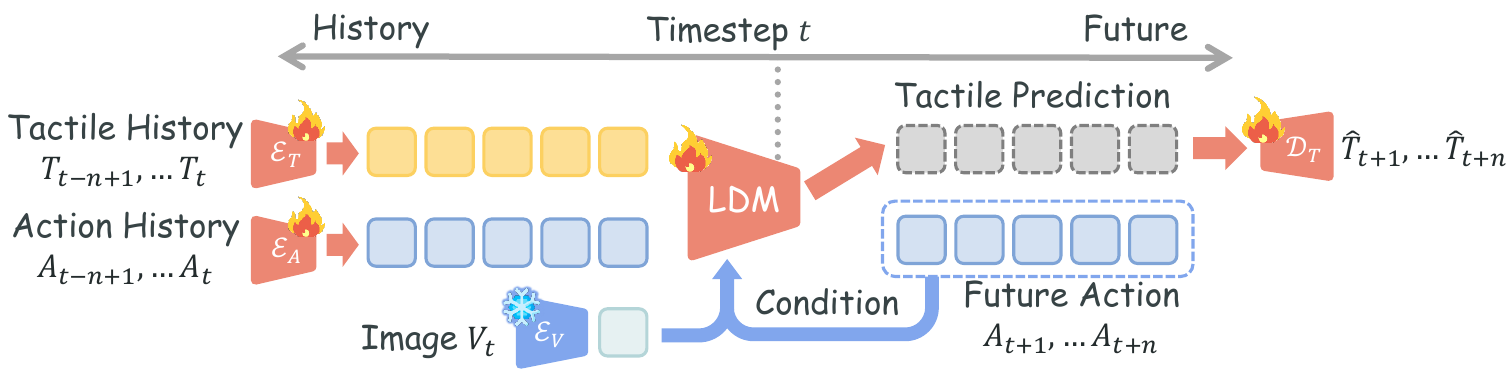}
\vspace{-5px}
\caption{The proposed tactile representation learning pipeline. The representation model $\mathcal{E}_T$ is learned during the temporal tactile prediction task. The history tactile and action features are fused and mapped as the prediction of future tactile features with a latent diffusion model (\texttt{LDM}). The future action and the current image are encoded as the condition of the denoising process.
}
\label{fig:pipeline}
\vspace{-5px}
\end{figure}

Given the exUMI hardware, we present an action-aware, and task-agnostic tactile representation learning framework that addresses the challenges of contact dynamics modeling in manipulation. We propose \textbf{T}actile \textbf{P}redictive \textbf{P}retraining (\textbf{TPP}), formulating tactile representation learning as a conditional future prediction: 
$p_\theta\left(\mathbf{T}_{t+1:t+n} | \mathcal{E}_T(\mathbf{T}_{t-n+1:t}), \mathcal{E}_V(\mathbf{V}_t), \mathcal{E}_A(\mathbf{A}_{t-n+1:t+n})\right)$,
The tactile encoder $\mathcal{E}_T$ would learn an informative representation involving tactile dynamics in the predictive pretraining.
The predictive model is learned from our 1~M human play data.
We adopt the diffusion model and masked autoencoder structure following \cite{uva} (Fig.~\ref{fig:pipeline}), with the following components:

\textbf{Multimodal Encoding}. 
We employ the VAE model as the encoder and decoder of the tactile modality ($\mathcal{E}_T,\mathcal{D}_T$), and each tactile image is patchified and converted into a sequence of embeddings. Different from UVA~\cite{uva}, the VAE model is learnable for tactile representation learning. 

\textbf{Tactile Prediction}.
Following \citet{uva}, we apply random masking to the history tactile patch embeddings and action features and fuse the two modalities with a transformer. The $n$ history latents are forwarded to a latent diffusion model to predict the latents of future tactile signals, where the embeddings of future action $A_{t+1:t+n}$ and current RGB image $V_t$ serve as the condition. The 
\begin{wrapfigure}{r}{0.38\linewidth}
    \centering
    \resizebox{\linewidth}{!}{
    \begin{tabular}{ccccc}
    \toprule
    \makecell{\bf Tactile \\ \bf History} & \makecell{\bf Action \\ \bf Input}   & \makecell{\bf RGB  \\ \bf Image}  & \makecell{\bf MSE \\ \bf Error}  \\
    \midrule
    \xmark  & \cmark &  \cmark &  0.0298     \\
    \cmark  & \xmark &  \xmark &  0.0132     \\
    \cmark  & \xmark &  \cmark &  0.0125     \\
    \cmark  & \cmark &  \xmark &  0.0117     \\
    \midrule
    \cmark  & \cmark &  \cmark &  \textbf{0.0099}     \\
    \bottomrule
    \end{tabular}
    }
    \captionof{table}{Tactile predictive pretraining with different input settings.}
    \vspace{-20px}
    \label{tab:tpp-mse}
\end{wrapfigure}
tactile image is reconstructed by the VAE decoder.
The predictive model is constrained by hybrid losses: 
(1) $\mathcal{L}_{diff}$: the regular diffusion loss between the predicted and actual noise perturbations.
(2) $\mathcal{L}_{recon}$: the reconstruction MSE loss between reconstructed and original tactile images.

\textbf{Policy Learning}.
After learning the predictive representation, we \textit{freeze} the tactile encoder $\mathcal{E}_T$ for all downstream policies. We learn the multimodal policy with common imitation learning.
The rich tactile dynamics knowledge has been encoded in the tactile model; hence, our approach could generate a more robust representation and alleviate the issues of data scarcity and low diversity, avoiding the overfitting in low-shot learning scenarios for tactile-aware tasks.
Note that although the pretraining process requires dense computational resources, the pretrained tactile model could be seamlessly adopted in all the downstream policy learning tasks without further finetuning.

\section{Experiment}
\label{sec:exp}

\subsection{Tactile Predictive Pretraining}

We first pretrain the TPP model on our collected large-scale dataset with a total of over 1M frames.

\textbf{Implementation Details.}
exUMI collects two tactile images on the two sides. We convert the images to a calibrated grayscale image following~\citet{9dtact}, and extract the convex and concave pixel maps and stack them as a 3-channel image for a richer representation of tactile contacts. 
We train the TPP model on 4 NVIDIA H100 GPUs for 120 hours. Refer to Appendix~\ref{app:sec:details} for details.

\textbf{Results.}
We compare the prediction MSE error on the val set in Tab.~\ref{tab:tpp-mse}, showing that multimodal conditioning of visual and action sequences could reduce the tactile prediction error. Among all the settings, our action-aware prediction achieved the best performance, indicating that the policy implicitly learns the forward tactile dynamics with full consideration of the action sequence.

\begin{figure}[t]
    \centering
    \begin{minipage}[c]{0.5\linewidth}
        \centering
        \includegraphics[width=\linewidth]{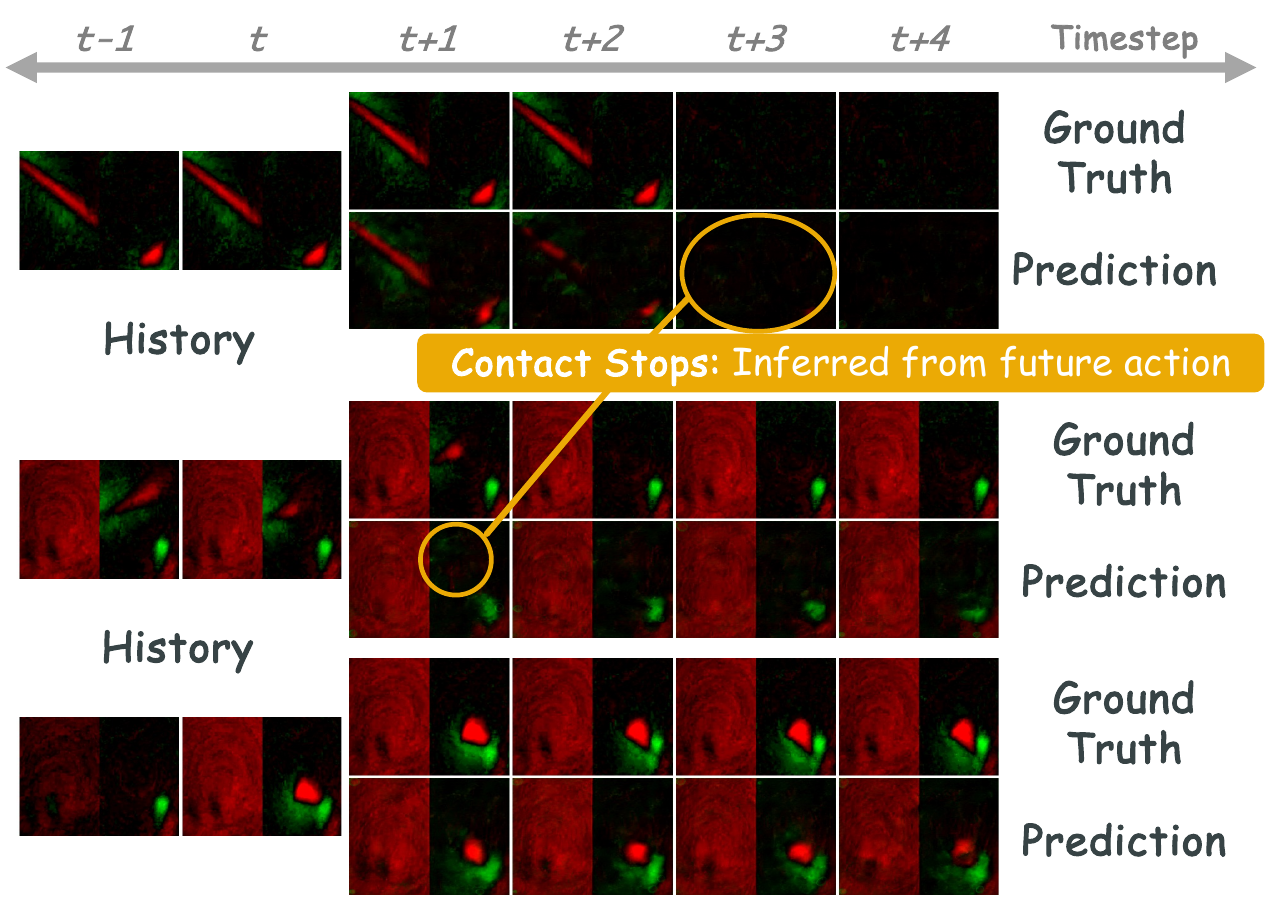}
        \caption{Examples of temporal tactile prediction of TPP on the validation set.}
        \label{fig:tactile-prediction}
    \end{minipage}
    \hfill
    \begin{minipage}[c]{0.48\linewidth}
        \centering
        \resizebox{\linewidth}{!}{
        \begin{tabular}{lccc}
        \toprule
        \textbf{Task} & \makecell{\bf Number of \\ \bf Demos}   & \makecell{\bf Data Collection \\ \bf Duration~$^\dagger$}  & \makecell{\bf Success \\ \bf Rate}  \\
        \midrule
        Pick Cube     & 204  &  42 min  &   \bf 85\%    \\
        Pick Carrot   & 135  &  31 min  &   \bf 80\%    \\
        Pick Broccoli & 170  &  47 min  &   \bf 60\%    \\
        Stack Cube    & 201  &  26 min  &   \bf 60\%    \\
        Insert Pen    & 139  &  60 min  &   \bf 65\%    \\
        \midrule
        Put Ball      & 181  &  66 min  &   \bf 70\%    \\
        Open Bottle   & 270  &  79 min  &   \bf 20\%    \\
        Pull Drawer   & 202  &  70 min  &   \bf 40\%    \\
        Peg in Hole   & 163  &  56 min  &   \bf 50\%    \\
        \bottomrule
        \end{tabular}
        }
        \captionof{table}{Data collection efficiency and policy performance on non-tactile tasks. $^\dagger$: total collection time of \textbf{all} demonstrations, including the environment resetting.}
        \label{tab:non-tactile-tasks}
    \end{minipage}
    \vspace{-10px}
\end{figure}

In Fig.~\ref{fig:tactile-prediction}. We visualize the tactile signals on the validation set (unseen data) following~\citet{9dtact}, where red and green represent the concave and convex areas. Our model manifests to learn clues from the action sequence. For example, in the first case, the model infers from the future action that the tactile contact will cease after two frames, which is \textit{impossible} for models without the reasoning of action-informed tactile dynamics. 
Thus, our action-aware predictive pretraining enables robust tactile representation for future tactile-aware policy learning.

\subsection{Experiment Settings for Imitation Learning}


\textbf{Environment Settings.}
We evaluate our learning system on a Flexiv Rizon 4 robot arm with a Flexiv Grav adaptive gripper, and use a GoPro camera as the only visual input.
We adopt diffusion policy~\cite{dp} with a ViT image backbone model, and direct feature concatenation for multimodal inputs.
We evaluate our policy for 20 trials.

\textbf{Task Settings.}
Our evaluation covers \underline{regular manipulation tasks}:
(1) {\it Pick cube / carrot / broccoli}: pick up and place it into a container;
(2) {\it Insert pen}: move a pen to another cup;
(3) {\it Stack cubes}: stack a small cube on top of another.

For \underline{tactile-aware} policy learning, we evaluate on more complex tasks:
(1) {\it Put Ball}: pick up a soft ball and place it in a cup.
(2) {\it Open Bottle}: rotate the bottle cap until it is fully unscrewed. 
(3) {\it Pull Drawer}: pull out one drawer, which is either empty (``Empty'') or contains a random amount of stones (``Random''), requiring tactile clues to determine the pulling direction.
(4) {\it Peg in hole}: insert a block into a slot, requiring a precision and force-aware adjustment. We split the task into ``Grasp'' and ``Insert'' stages.
Please refer to Appendix~\ref{app:sec:details} for more details.

\textbf{Demonstration Collection.}
We collect 100 to 200 demonstrations with exUMI for robot teaching, and the total data collection time is shown in Tab.~\ref{tab:non-tactile-tasks}. 
Note that the time duration includes the task. With its efficiency and tactile feedback, an expert could complete the data collection within half an hour.
And our system has a nearly 100\% effective data ratio. In comparison, users should spend \textbf{50\%} or more data collection time with the regular UMI system for the same amount of valid data.

\subsection{Real World Evaluation Result}

\textbf{Non-tactile Imitation Learning}.
We first train a vision-only diffusion policy to evaluate the data collection quality of exUMI (Tab.~\ref{tab:non-tactile-tasks}). The policy achieves decent performance across tasks, demonstrating both sufficient spatial demonstration quality and its capability for visual imitation learning. 
Our policy achieves an over 80\% success rate on simple pick-and-place tasks, indicating sufficient spatial demonstration quality. The success rate of ``pick broccoli'' task slightly drops to 60\% due to the slippery surface and irregular geometry.
Tasks involving precision manipulation (``stack'' and ``insert'') are more challenging, but we still reach over 60\% due to the accurate robot proprioception of exUMI.
Complex tasks requiring reasoning with tactile feedback show performance degradation. ``Pull drawer'' achieves merely a 40\% success rate, and the majority of failures occur when the policy applies excessive force in incorrect directions. For ``peg in hole'', the policy usually fails to align the peg merely on visual input. These tasks require further tactile robot learning.


\begin{table}[t]
    \centering
    \resizebox{0.95\linewidth}{!}{
    \begin{tabular}{l|c|c|cc|cc}
    \toprule
    \multirow{ 2}{*}{\bf Input Representation} & \multirow{ 2}{*}{\bf Put Ball} & \multirow{ 2}{*}{\bf Open Bottle} & \multicolumn{2}{c|}{\bf Pull Drawer} & \multicolumn{2}{c}{\bf Peg in Hole} \\
    &   &  &  Empty  &  Random  & Grasp & Insert  \\
    \midrule
    Vision Only                     &     70\% & 20\%  &   100\%  &  40\%     & 100\%  &      50\%    \\
    Vision \& Tactile               &     70\% & 50\%  &   100\%  &  50\%     & 100\%  &      60\%    \\
    Vision \& Tactile w/ TPP (Ours) & \bf 85\% & \bf 60\%  &   100\%  & \bf 95\%  & 100\%  &  \bf 80\%   \\
    \bottomrule
    \end{tabular}
    }
    \caption{Real world evaluation of tactile-aware policies on complex tasks.}
    \label{tab:tactile-tasks}
    \vspace{-14px}
\end{table}

\begin{figure}[t]
    \centering
    \includegraphics[width=\linewidth]{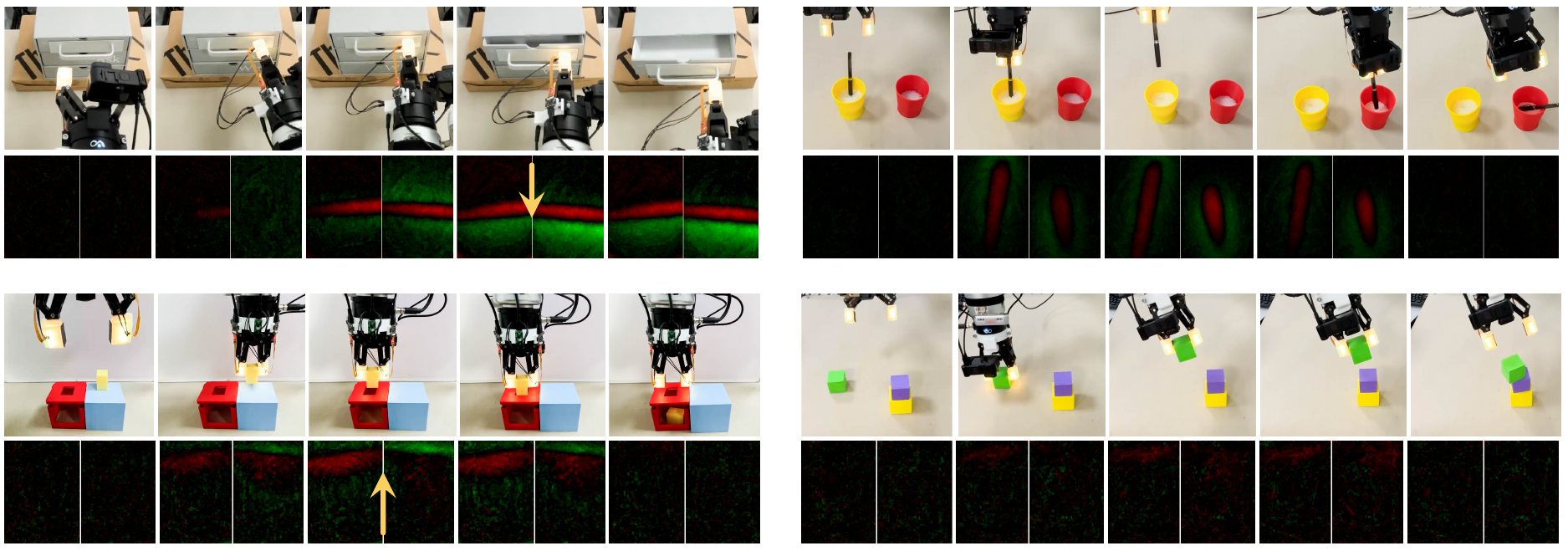}
    \caption{Real world rollout of tactile-aware policy. Yellow arrows indicate the tangent force, which points from red area (concave due to pressure) to green area (convex bump of silicone gel).}
    \vspace{-8px}
    \label{fig:tactile-rollout}
\end{figure}

\textbf{Tactile-aware Imitation Learning}.
We augment our policy model with tactile sensing and evaluate across contact-sensitive tasks (Tab.~\ref{tab:tactile-tasks} and visualized in Fig.~\ref{fig:tactile-rollout}). While both vision-only and tactile-aware policies achieve a high success rate on simple settings, tactile integration yields significant improvements, particularly in harder stages (\eg drawer pulling ).
For the ``put ball'' and ``rotate bottle'' tasks, TPP brings $>$15\% performance gain over vanilla tactile policy. And though tactile policy is only comparable to vision policy on ``put ball'' task, we empirically observe that tactile-aware policy slightly adjusts the grasp pose until accurately holding the center of the ball.
For the ``pull drawer'' and ``peg in hole'' tasks, all policies with or without tactile sensing achieve a high success rate of 100\% at non-tactile stages (pulling the empty drawer or simply grasping the object). But for the force-sensitive stages, tactile-aware policies continuously bring improvement, and our TPP boosts the performance to over 100\% and 80\%, respectively. For these tasks, tactile clues play an important role in the trajectory planning. As shown in Fig.~\ref{fig:tactile-rollout}, for the ``pull drawer'' task, the tactile signals inform the grasp pose of the handle (red area), which is critical for the selection of pulling direction.
This indicates that the insertion success depends critically on post-grasp tactile feedback.


These results validate that effective tactile integration requires both careful tactile feature engineering (via pretraining). The performance gaps between our method and baselines highlight the importance of learning the temporal dynamics of tactile for real-world manipulation.

\section{Conclusion}
\label{sec:conclusion}

In this work, we propose a co-design of the exUMI hardware with its reliable proprioception and scalable tactile sensing, and the TPP framework, which learns tactile features by predictive proxy task. Our system achieves significant performance gain on complex tactile-aware tasks, highlighting the importance of grounding human-style contact dynamics reasoning in physical interaction.

\clearpage

\section{Limitations}

While our system provides a user-friendly interface for tactile-aware robot data collection, several critical challenges remain to be addressed in future research.

\textbf{Hardware Limitations.} (1) Although our AR headset-based motion capture system demonstrates robustness, user feedback highlights two ergonomic concerns of thermal discomfort and neck strain. While we considered alternative tracking solutions (\eg, dedicated motion capture trackers like HTC Vive Tracker), these trackers usually rely on external base stations for more accurate tracking, which conflicts with our design goal of maintaining portability for in-situ AR-assisted data acquisition. Future work could explore ergonomic upgrades like neck supports.
(2) The durability and consistency have always been critical concerns of tactile sensors. We adopt the 9DTact sensor for a low-cost tactile solution. While we have notably improved tactile sensor consistency to 9DTact, further enhancements remain possible.

\textbf{Algorithm Limitations.}
Our predictive framework for learning interaction representations faces inherent constraints. The interaction and movement information is limited due to low action dimension and limited camera angle, resulting in imperfect tactile prediction performance. We plan to address these by integrating force-torque measurements and multi-view vision inputs in the future.

%





\bibliography{main}  

\begin{thebibliography}{53}
\providecommand{\natexlab}[1]{#1}
\providecommand{\url}[1]{\texttt{#1}}
\expandafter\ifx\csname urlstyle\endcsname\relax
  \providecommand{\doi}[1]{doi: #1}\else
  \providecommand{\doi}{doi: \begingroup \urlstyle{rm}\Url}\fi

\bibitem[Chi et~al.(2024)Chi, Xu, Pan, Cousineau, Burchfiel, Feng, Tedrake, and Song]{umi}
C.~Chi, Z.~Xu, C.~Pan, E.~Cousineau, B.~Burchfiel, S.~Feng, R.~Tedrake, and S.~Song.
\newblock Universal manipulation interface: In-the-wild robot teaching without in-the-wild robots.
\newblock \emph{arXiv preprint arXiv:2402.10329}, 2024.

\bibitem[Zhu et~al.(2023)Zhu, Joshi, Stone, and Zhu]{zhu2023viola}
Y.~Zhu, A.~Joshi, P.~Stone, and Y.~Zhu.
\newblock Viola: Imitation learning for vision-based manipulation with object proposal priors.
\newblock In \emph{Conference on Robot Learning}, pages 1199--1210. PMLR, 2023.

\bibitem[Mandlekar et~al.(2018)Mandlekar, Zhu, Garg, Booher, Spero, Tung, Gao, Emmons, Gupta, Orbay, et~al.]{mandlekar2018roboturk}
A.~Mandlekar, Y.~Zhu, A.~Garg, J.~Booher, M.~Spero, A.~Tung, J.~Gao, J.~Emmons, A.~Gupta, E.~Orbay, et~al.
\newblock Roboturk: A crowdsourcing platform for robotic skill learning through imitation.
\newblock In \emph{Conference on Robot Learning}, pages 879--893. PMLR, 2018.

\bibitem[Luo et~al.(2024)Luo, Peng, Lv, Hong, Driggs-Campbell, Lu, and Li]{luo2024human}
S.~Luo, Q.~Peng, J.~Lv, K.~Hong, K.~R. Driggs-Campbell, C.~Lu, and Y.-L. Li.
\newblock Human-agent joint learning for efficient robot manipulation skill acquisition.
\newblock \emph{arXiv preprint arXiv:2407.00299}, 2024.

\bibitem[Shao et~al.(2021)Shao, Migimatsu, Zhang, Yang, and Bohg]{shao2021concept2robot}
L.~Shao, T.~Migimatsu, Q.~Zhang, K.~Yang, and J.~Bohg.
\newblock Concept2robot: Learning manipulation concepts from instructions and human demonstrations.
\newblock \emph{The International Journal of Robotics Research}, 40\penalty0 (12-14):\penalty0 1419--1434, 2021.

\bibitem[Chen et~al.(2021)Chen, Nair, and Finn]{chen2021learning}
A.~S. Chen, S.~Nair, and C.~Finn.
\newblock Learning generalizable robotic reward functions from" in-the-wild" human videos.
\newblock \emph{arXiv preprint arXiv:2103.16817}, 2021.

\bibitem[Bahl et~al.(2022)Bahl, Gupta, and Pathak]{bahl2022human}
S.~Bahl, A.~Gupta, and D.~Pathak.
\newblock Human-to-robot imitation in the wild.
\newblock \emph{arXiv preprint arXiv:2207.09450}, 2022.

\bibitem[Simeonov et~al.(2022)Simeonov, Du, Tagliasacchi, Tenenbaum, Rodriguez, Agrawal, and Sitzmann]{simeonov2022neural}
A.~Simeonov, Y.~Du, A.~Tagliasacchi, J.~B. Tenenbaum, A.~Rodriguez, P.~Agrawal, and V.~Sitzmann.
\newblock Neural descriptor fields: Se (3)-equivariant object representations for manipulation.
\newblock In \emph{2022 International Conference on Robotics and Automation (ICRA)}, pages 6394--6400. IEEE, 2022.

\bibitem[Schmeckpeper et~al.(2020)Schmeckpeper, Xie, Rybkin, Tian, Daniilidis, Levine, and Finn]{schmeckpeper2020learning}
K.~Schmeckpeper, A.~Xie, O.~Rybkin, S.~Tian, K.~Daniilidis, S.~Levine, and C.~Finn.
\newblock Learning predictive models from observation and interaction.
\newblock In \emph{European Conference on Computer Vision}, pages 708--725. Springer, 2020.

\bibitem[Qin et~al.(2022)Qin, Wu, Liu, Jiang, Yang, Fu, and Wang]{qin2022dexmv}
Y.~Qin, Y.-H. Wu, S.~Liu, H.~Jiang, R.~Yang, Y.~Fu, and X.~Wang.
\newblock Dexmv: Imitation learning for dexterous manipulation from human videos.
\newblock In \emph{European Conference on Computer Vision}, pages 570--587. Springer, 2022.

\bibitem[Xiao et~al.(2022)Xiao, Radosavovic, Darrell, and Malik]{xiao2022masked}
T.~Xiao, I.~Radosavovic, T.~Darrell, and J.~Malik.
\newblock Masked visual pre-training for motor control.
\newblock \emph{arXiv preprint arXiv:2203.06173}, 2022.

\bibitem[Fu et~al.(2024)Fu, Datta, Huang, Panitch, Drake, Ortiz, Mukadam, Lambeta, Calandra, and Goldberg]{fu2024touch}
L.~Fu, G.~Datta, H.~Huang, W.~C.-H. Panitch, J.~Drake, J.~Ortiz, M.~Mukadam, M.~Lambeta, R.~Calandra, and K.~Goldberg.
\newblock A touch, vision, and language dataset for multimodal alignment.
\newblock \emph{arXiv preprint arXiv:2402.13232}, 2024.

\bibitem[Su et~al.(2024)Su, Jia, Qin, Zhou, Macaluso, Huang, and Wang]{10611113}
E.~Su, C.~Jia, Y.~Qin, W.~Zhou, A.~Macaluso, B.~Huang, and X.~Wang.
\newblock Sim2real manipulation on unknown objects with tactile-based reinforcement learning.
\newblock In \emph{2024 IEEE International Conference on Robotics and Automation (ICRA)}, pages 9234--9241, 2024.
\newblock \doi{10.1109/ICRA57147.2024.10611113}.

\bibitem[Lee et~al.(2024)Lee, Qin, Wang, and Lim]{Lee_2024}
K.-W. Lee, Y.~Qin, X.~Wang, and S.-C. Lim.
\newblock Dextouch: Learning to seek and manipulate objects with tactile dexterity.
\newblock \emph{IEEE Robotics and Automation Letters}, 9\penalty0 (12):\penalty0 10772–10779, Dec. 2024.
\newblock ISSN 2377-3774.
\newblock \doi{10.1109/lra.2024.3478571}.
\newblock URL \url{http://dx.doi.org/10.1109/LRA.2024.3478571}.

\bibitem[Rodriguez et~al.(2024)Rodriguez, Dou, van~den Bogert, Oller, So, Owens, and Fazeli]{rodriguez2024contrastivetouchtotouchpretraining}
S.~Rodriguez, Y.~Dou, W.~van~den Bogert, M.~Oller, K.~So, A.~Owens, and N.~Fazeli.
\newblock Contrastive touch-to-touch pretraining, 2024.
\newblock URL \url{https://arxiv.org/abs/2410.11834}.

\bibitem[Guzey et~al.(2023)Guzey, Evans, Chintala, and Pinto]{guzey2023dexteritytouchselfsupervisedpretraining}
I.~Guzey, B.~Evans, S.~Chintala, and L.~Pinto.
\newblock Dexterity from touch: Self-supervised pre-training of tactile representations with robotic play, 2023.
\newblock URL \url{https://arxiv.org/abs/2303.12076}.

\bibitem[Yu et~al.(2025)Yu, Han, Wang, Saxena, Xu, and Zhao]{yu2025mimictouchleveragingmultimodalhuman}
K.~Yu, Y.~Han, Q.~Wang, V.~Saxena, D.~Xu, and Y.~Zhao.
\newblock Mimictouch: Leveraging multi-modal human tactile demonstrations for contact-rich manipulation, 2025.
\newblock URL \url{https://arxiv.org/abs/2310.16917}.

\bibitem[Wu et~al.(2025)Wu, Li, Zhang, Wu, and Dong]{wu2025canonicalrepresentationforcebasedpretraining}
T.~Wu, J.~Li, J.~Zhang, M.~Wu, and H.~Dong.
\newblock Canonical representation and force-based pretraining of 3d tactile for dexterous visuo-tactile policy learning, 2025.
\newblock URL \url{https://arxiv.org/abs/2409.17549}.

\bibitem[Romero et~al.(2024)Romero, Fang, Agrawal, and Adelson]{10802778}
B.~Romero, H.-S. Fang, P.~Agrawal, and E.~Adelson.
\newblock Eyesight hand: Design of a fully-actuated dexterous robot hand with integrated vision-based tactile sensors and compliant actuation.
\newblock In \emph{2024 IEEE/RSJ International Conference on Intelligent Robots and Systems (IROS)}, pages 1853--1860, 2024.
\newblock \doi{10.1109/IROS58592.2024.10802778}.

\bibitem[Lin et~al.(2024)Lin, Corcodel, and Zhao]{10610567}
H.~Lin, R.~Corcodel, and D.~Zhao.
\newblock Generalize by touching: Tactile ensemble skill transfer for robotic furniture assembly.
\newblock In \emph{2024 IEEE International Conference on Robotics and Automation (ICRA)}, pages 9227--9233, 2024.
\newblock \doi{10.1109/ICRA57147.2024.10610567}.

\bibitem[Pattabiraman et~al.(2024)Pattabiraman, Cao, Haldar, Pinto, and Bhirangi]{pattabiraman2024learningprecisecontactrichmanipulation}
V.~Pattabiraman, Y.~Cao, S.~Haldar, L.~Pinto, and R.~Bhirangi.
\newblock Learning precise, contact-rich manipulation through uncalibrated tactile skins, 2024.
\newblock URL \url{https://arxiv.org/abs/2410.17246}.

\bibitem[Lin et~al.(2023)Lin, Zhang, Xu, Wu, and Xu]{9dtact}
C.~Lin, H.~Zhang, J.~Xu, L.~Wu, and H.~Xu.
\newblock 9dtact: A compact vision-based tactile sensor for accurate 3d shape reconstruction and generalizable 6d force estimation.
\newblock \emph{IEEE Robotics and Automation Letters}, 2023.

\bibitem[Fan et~al.(2023)Fan, Guo, Feng, Lin, Wang, Liang, Garrad, Rossiter, Zhang, Lepora, et~al.]{fan2023digital}
W.~Fan, X.~Guo, E.~Feng, J.~Lin, Y.~Wang, J.~Liang, M.~Garrad, J.~Rossiter, Z.~Zhang, N.~Lepora, et~al.
\newblock Digital twin-driven mixed reality framework for immersive teleoperation with haptic rendering.
\newblock \emph{IEEE Robotics and Automation Letters}, 2023.

\bibitem[Chi et~al.(2023)Chi, Xu, Feng, Cousineau, Du, Burchfiel, Tedrake, and Song]{chi2023diffusion}
C.~Chi, Z.~Xu, S.~Feng, E.~Cousineau, Y.~Du, B.~Burchfiel, R.~Tedrake, and S.~Song.
\newblock Diffusion policy: Visuomotor policy learning via action diffusion.
\newblock \emph{The International Journal of Robotics Research}, page 02783649241273668, 2023.

\bibitem[Seo et~al.(2023)Seo, Han, Sim, Bang, Gonzalez, Sentis, and Zhu]{seo2023deep}
M.~Seo, S.~Han, K.~Sim, S.~H. Bang, C.~Gonzalez, L.~Sentis, and Y.~Zhu.
\newblock Deep imitation learning for humanoid loco-manipulation through human teleoperation.
\newblock In \emph{2023 IEEE-RAS 22nd International Conference on Humanoid Robots (Humanoids)}, pages 1--8. IEEE, 2023.

\bibitem[Ma et~al.(2022)Ma, Sodhani, Jayaraman, Bastani, Kumar, and Zhang]{ma2022vip}
Y.~J. Ma, S.~Sodhani, D.~Jayaraman, O.~Bastani, V.~Kumar, and A.~Zhang.
\newblock Vip: Towards universal visual reward and representation via value-implicit pre-training.
\newblock \emph{arXiv preprint arXiv:2210.00030}, 2022.

\bibitem[Bahl et~al.(2023)Bahl, Mendonca, Chen, Jain, and Pathak]{bahl2023affordances}
S.~Bahl, R.~Mendonca, L.~Chen, U.~Jain, and D.~Pathak.
\newblock Affordances from human videos as a versatile representation for robotics.
\newblock In \emph{Proceedings of the IEEE/CVF Conference on Computer Vision and Pattern Recognition}, pages 13778--13790, 2023.

\bibitem[Pan et~al.(2023)Pan, Okorn, Zhang, Eisner, and Held]{pan2023tax}
C.~Pan, B.~Okorn, H.~Zhang, B.~Eisner, and D.~Held.
\newblock Tax-pose: Task-specific cross-pose estimation for robot manipulation.
\newblock In \emph{Conference on Robot Learning}, pages 1783--1792. PMLR, 2023.

\bibitem[Shen et~al.(2023)Shen, Yang, Yu, Wong, Kaelbling, and Isola]{shen2023distilled}
W.~Shen, G.~Yang, A.~Yu, J.~Wong, L.~P. Kaelbling, and P.~Isola.
\newblock Distilled feature fields enable few-shot language-guided manipulation.
\newblock \emph{arXiv preprint arXiv:2308.07931}, 2023.

\bibitem[Nair et~al.(2022)Nair, Rajeswaran, Kumar, Finn, and Gupta]{nair2022r3m}
S.~Nair, A.~Rajeswaran, V.~Kumar, C.~Finn, and A.~Gupta.
\newblock R3m: A universal visual representation for robot manipulation.
\newblock \emph{arXiv preprint arXiv:2203.12601}, 2022.

\bibitem[Fang et~al.(2024)Fang, Fang, Wang, Ren, Chen, Zhang, Wang, and Lu]{fang2024airexo}
H.~Fang, H.-S. Fang, Y.~Wang, J.~Ren, J.~Chen, R.~Zhang, W.~Wang, and C.~Lu.
\newblock Airexo: Low-cost exoskeletons for learning whole-arm manipulation in the wild.
\newblock In \emph{2024 IEEE International Conference on Robotics and Automation (ICRA)}, pages 15031--15038. IEEE, 2024.

\bibitem[Wu et~al.(2024)Wu, Wang, Guan, Jia, Liang, Song, Qu, Wang, Wang, Cao, et~al.]{wu2024fast}
Z.~Wu, T.~Wang, C.~Guan, Z.~Jia, S.~Liang, H.~Song, D.~Qu, D.~Wang, Z.~Wang, N.~Cao, et~al.
\newblock Fast-umi: A scalable and hardware-independent universal manipulation interface.
\newblock \emph{arXiv preprint arXiv:2409.19499}, 2024.

\bibitem[Liu et~al.(2024)Liu, Wang, Wang, Wang, and Lu]{liu2024forcemimic}
W.~Liu, J.~Wang, Y.~Wang, W.~Wang, and C.~Lu.
\newblock Forcemimic: Force-centric imitation learning with force-motion capture system for contact-rich manipulation.
\newblock \emph{arXiv preprint arXiv:2410.07554}, 2024.

\bibitem[Yuan et~al.(2024)Yuan, Che, Qin, Huang, Yin, Lee, Wu, Lim, and Wang]{10610532}
Y.~Yuan, H.~Che, Y.~Qin, B.~Huang, Z.-H. Yin, K.-W. Lee, Y.~Wu, S.-C. Lim, and X.~Wang.
\newblock Robot synesthesia: In-hand manipulation with visuotactile sensing.
\newblock In \emph{2024 IEEE International Conference on Robotics and Automation (ICRA)}, pages 6558--6565, 2024.
\newblock \doi{10.1109/ICRA57147.2024.10610532}.

\bibitem[Huang et~al.(2025)Huang, Wang, Yang, Luo, and Li]{huang20253dvitaclearningfinegrainedmanipulation}
B.~Huang, Y.~Wang, X.~Yang, Y.~Luo, and Y.~Li.
\newblock 3d-vitac: Learning fine-grained manipulation with visuo-tactile sensing, 2025.
\newblock URL \url{https://arxiv.org/abs/2410.24091}.

\bibitem[Suresh et~al.(2024)Suresh, Qi, Wu, Fan, Pineda, Lambeta, Malik, Kalakrishnan, Calandra, Kaess, Ortiz, and Mukadam]{doi:10.1126/scirobotics.adl0628}
S.~Suresh, H.~Qi, T.~Wu, T.~Fan, L.~Pineda, M.~Lambeta, J.~Malik, M.~Kalakrishnan, R.~Calandra, M.~Kaess, J.~Ortiz, and M.~Mukadam.
\newblock Neuralfeels with neural fields: Visuotactile perception for in-hand manipulation.
\newblock \emph{Science Robotics}, 9\penalty0 (96):\penalty0 eadl0628, 2024.
\newblock \doi{10.1126/scirobotics.adl0628}.
\newblock URL \url{https://www.science.org/doi/abs/10.1126/scirobotics.adl0628}.

\bibitem[Feng et~al.(2025)Feng, Hu, Xia, Gao, Shen, Sun, Fang, and Hu]{feng2025anytouch}
R.~Feng, J.~Hu, W.~Xia, T.~Gao, A.~Shen, Y.~Sun, B.~Fang, and D.~Hu.
\newblock Anytouch: Learning unified static-dynamic representation across multiple visuo-tactile sensors.
\newblock \emph{arXiv preprint arXiv:2502.12191}, 2025.

\bibitem[Zhao et~al.(2024)Zhao, Ma, Wang, and Adelson]{zhao2024transferable}
J.~Zhao, Y.~Ma, L.~Wang, and E.~H. Adelson.
\newblock Transferable tactile transformers for representation learning across diverse sensors and tasks.
\newblock \emph{arXiv preprint arXiv:2406.13640}, 2024.

\bibitem[Calandra et~al.(2018)Calandra, Owens, Jayaraman, Lin, Yuan, Malik, Adelson, and Levine]{MoreThanFeel}
R.~Calandra, A.~Owens, D.~Jayaraman, J.~Lin, W.~Yuan, J.~Malik, E.~H. Adelson, and S.~Levine.
\newblock More than a feeling: Learning to grasp and regrasp using vision and touch.
\newblock \emph{IEEE Robotics and Automation Letters}, 3\penalty0 (4):\penalty0 3300--3307, 2018.

\bibitem[Calandra et~al.(2017)Calandra, Owens, Upadhyaya, Yuan, Lin, Adelson, and Levine]{FeelSucc}
R.~Calandra, A.~Owens, M.~Upadhyaya, W.~Yuan, J.~Lin, E.~H. Adelson, and S.~Levine.
\newblock The feeling of success: Does touch sensing help predict grasp outcomes?
\newblock \emph{arXiv preprint arXiv:1710.05512}, 2017.

\bibitem[Kerr et~al.(2022)Kerr, Huang, Wilcox, Hoque, Ichnowski, Calandra, and Goldberg]{kerr2022self}
J.~Kerr, H.~Huang, A.~Wilcox, R.~Hoque, J.~Ichnowski, R.~Calandra, and K.~Goldberg.
\newblock Self-supervised visuo-tactile pretraining to locate and follow garment features.
\newblock \emph{arXiv preprint arXiv:2209.13042}, 2022.

\bibitem[Li et~al.(2019)Li, Zhu, Tedrake, and Torralba]{VisGel}
Y.~Li, J.-Y. Zhu, R.~Tedrake, and A.~Torralba.
\newblock Connecting touch and vision via cross-modal prediction.
\newblock In \emph{Proceedings of the IEEE/CVF Conference on Computer Vision and Pattern Recognition}, pages 10609--10618, 2019.

\bibitem[Clarke et~al.(2025)Clarke, Wistreich, Ze, and Wu]{clarke2025x}
S.~Clarke, S.~Wistreich, Y.~Ze, and J.~Wu.
\newblock X-capture: An open-source portable device for multi-sensory learning.
\newblock \emph{arXiv preprint arXiv:2504.02318}, 2025.

\bibitem[Yang et~al.(2022)Yang, Ma, Zhang, Zhu, Yuan, and Owens]{yang2022touchgolearninghumancollected}
F.~Yang, C.~Ma, J.~Zhang, J.~Zhu, W.~Yuan, and A.~Owens.
\newblock Touch and go: Learning from human-collected vision and touch, 2022.
\newblock URL \url{https://arxiv.org/abs/2211.12498}.

\bibitem[Chen et~al.(2024)Chen, Wang, Nguyen, Fei-Fei, and Liu]{chen2024arcap}
S.~Chen, C.~Wang, K.~Nguyen, L.~Fei-Fei, and C.~K. Liu.
\newblock Arcap: Collecting high-quality human demonstrations for robot learning with augmented reality feedback.
\newblock \emph{arXiv preprint arXiv:2410.08464}, 2024.

\bibitem[Li et~al.(2025)Li, Gao, Sadigh, and Song]{uva}
S.~Li, Y.~Gao, D.~Sadigh, and S.~Song.
\newblock Unified video action model.
\newblock \emph{arXiv preprint arXiv:2503.00200}, 2025.

\bibitem[Chi et~al.(2023)Chi, Xu, Feng, Cousineau, Du, Burchfiel, Tedrake, and Song]{dp}
C.~Chi, Z.~Xu, S.~Feng, E.~Cousineau, Y.~Du, B.~Burchfiel, R.~Tedrake, and S.~Song.
\newblock Diffusion policy: Visuomotor policy learning via action diffusion.
\newblock \emph{The International Journal of Robotics Research}, page 02783649241273668, 2023.

\bibitem[Higuera et~al.(2024)Higuera, Sharma, Bodduluri, Fan, Lancaster, Kalakrishnan, Kaess, Boots, Lambeta, Wu, et~al.]{higuera2024sparsh}
C.~Higuera, A.~Sharma, C.~K. Bodduluri, T.~Fan, P.~Lancaster, M.~Kalakrishnan, M.~Kaess, B.~Boots, M.~Lambeta, T.~Wu, et~al.
\newblock Sparsh: Self-supervised touch representations for vision-based tactile sensing.
\newblock \emph{arXiv preprint arXiv:2410.24090}, 2024.

\bibitem[Xu et~al.(2025)Xu, Uppuluri, Zhang, Fitch, Crandall, Shou, Wang, and She]{xu2025unit}
Z.~Xu, R.~Uppuluri, X.~Zhang, C.~Fitch, P.~G. Crandall, W.~Shou, D.~Wang, and Y.~She.
\newblock Unit: Data efficient tactile representation with generalization to unseen objects.
\newblock \emph{IEEE Robotics and Automation Letters}, 2025.

\bibitem[Burka(2018)]{burka2018inst}
A.~L. Burka.
\newblock \emph{Instrumentation, data, and algorithms for visually understanding haptic surface properties}.
\newblock PhD thesis, University of Pennsylvania, 2018.

\bibitem[Gao et~al.(2023)Gao, Dou, Li, Agarwal, Bohg, Li, Fei-Fei, and Wu]{objfolder-r}
R.~Gao, Y.~Dou, H.~Li, T.~Agarwal, J.~Bohg, Y.~Li, L.~Fei-Fei, and J.~Wu.
\newblock The objectfolder benchmark: Multisensory learning with neural and real objects.
\newblock In \emph{Proceedings of the IEEE/CVF Conference on Computer Vision and Pattern Recognition}, pages 17276--17286, 2023.

\bibitem[Rodriguez et~al.(2024)Rodriguez, Dou, Oller, Owens, and Fazeli]{touch2touch}
S.~Rodriguez, Y.~Dou, M.~Oller, A.~Owens, and N.~Fazeli.
\newblock Touch2touch: Cross-modal tactile generation for object manipulation.
\newblock \emph{arXiv preprint arXiv:2409.08269}, 2024.

\bibitem[Yu et~al.(2023)Yu, Han, Wang, Saxena, Xu, and Zhao]{yu2023mimictouch}
K.~Yu, Y.~Han, Q.~Wang, V.~Saxena, D.~Xu, and Y.~Zhao.
\newblock Mimictouch: Leveraging multi-modal human tactile demonstrations for contact-rich manipulation.
\newblock \emph{arXiv preprint arXiv:2310.16917}, 2023.

\end{thebibliography}

\clearpage

\textbf{\LARGE Appendix}

\begin{appendix}

\section{Details of exUMI Hardware Design}
\label{sec:hardware-detail}

We gave a brief introduction to the hardware and algorithms for exUMI due to the page limit. Below are the details of our system.

\subsection{Hardware}

\begin{figure}[h]
    \centering
    \includegraphics[width=0.8\linewidth]{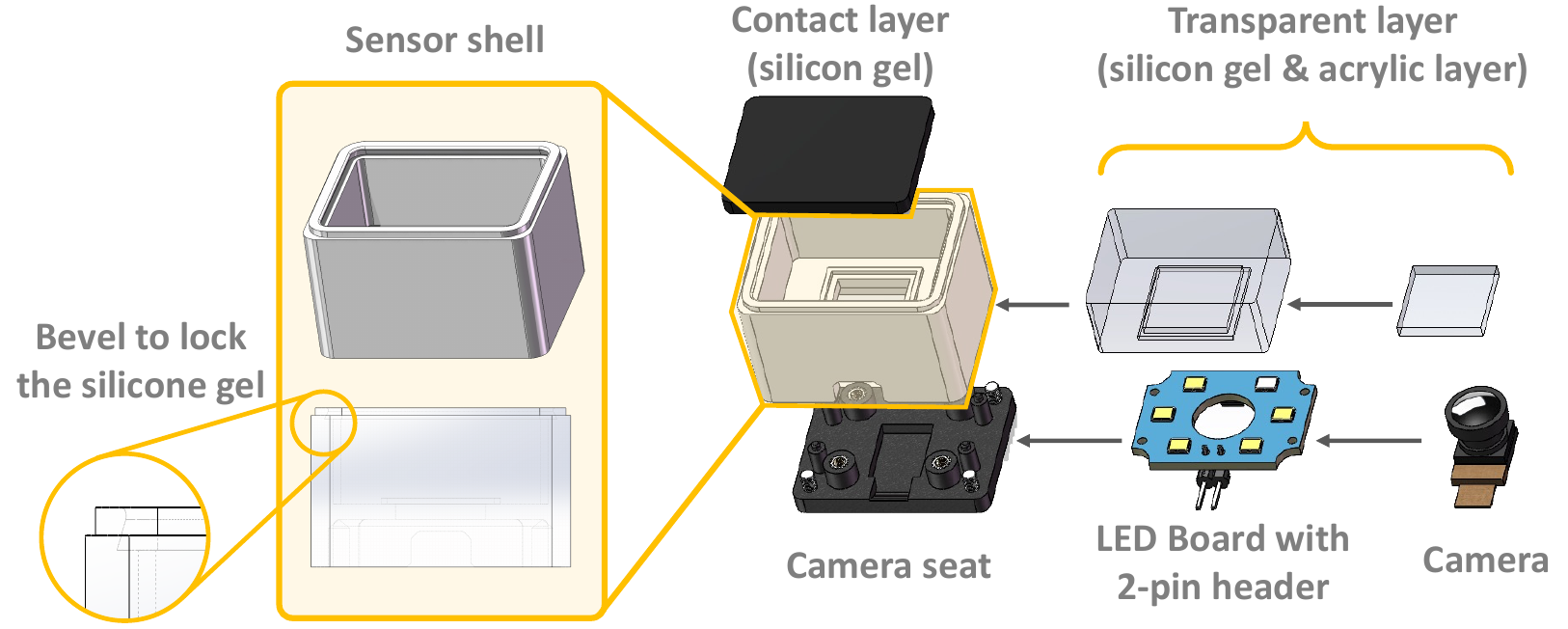}
    \caption{An exploded view of the enhanced tactile sensor design. We enhance the 9DTact for stability and quality control. We add an bevel to the sensor shell to secure the black silicon gel and prevent it from detaching.}
    \label{fig:9dtact}
\end{figure}

\begin{figure}[t]
    \centering
    \begin{minipage}[c]{0.42\linewidth}
        \centering
        \includegraphics[width=\linewidth]{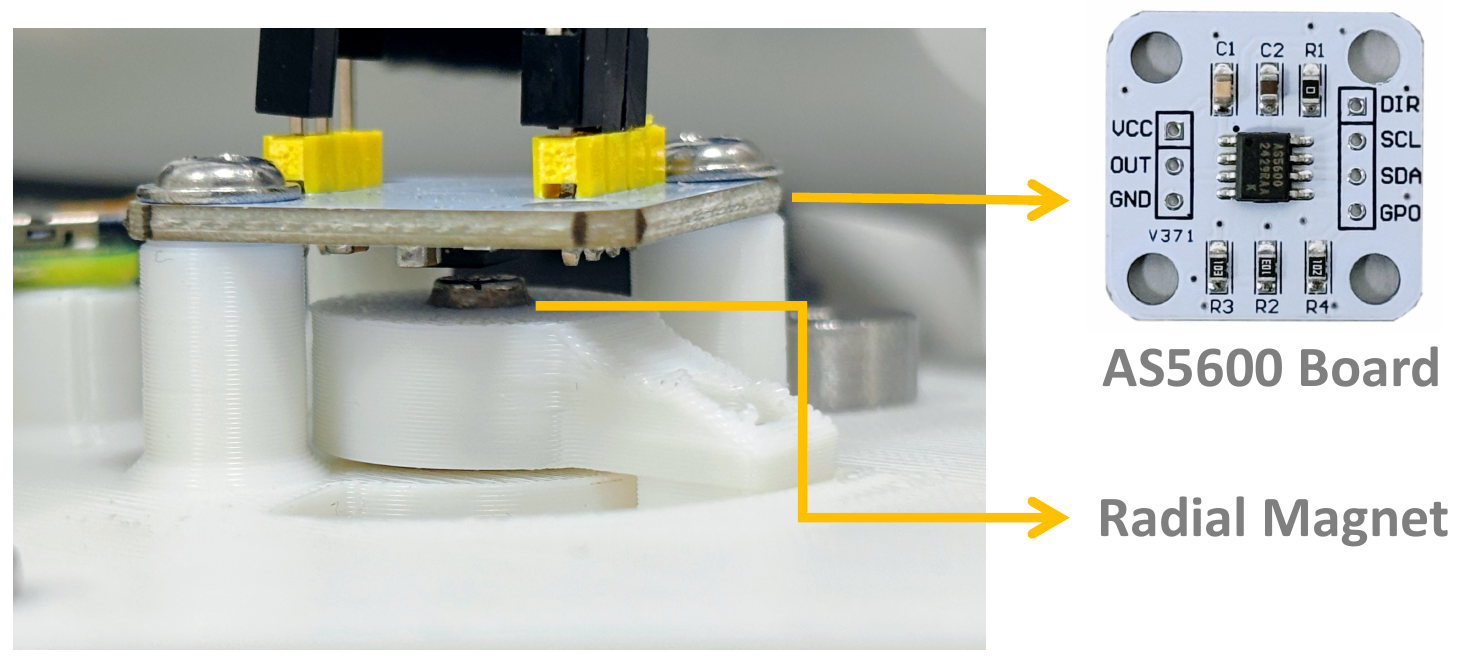}
        \caption{Detailed view of AS5600 sensor on exUMI.}
        \label{fig:as5600}
    \end{minipage}
    \hfill
    \begin{minipage}[c]{0.55\linewidth}
        \resizebox{\linewidth}{!}{%
            \begin{tabular}{l|l}
            \toprule
            Component               & Base Cost (\$)    \\
            \midrule
            GoPro 11 + Accessories  & 298    \\
            Meta Quest VR Headset   & 299    \\
            Orange Pi 3B            & 35     \\
            AS5600 Magnetic Encoder & 1      \\
            3D Printed Parts        & 15     \\
            Visuo-Tactile Sensors   & 30     \\
            Misc. (Power Bank, Cables/Screws/Nuts) & 20  \\
            \midrule
            \textbf{Total Cost of exUMI} & \textbf{698}  \\
            \bottomrule
            \end{tabular}
        }
        \captionof{table}{Bill of materials (BOM) of exUMI. \label{tab:bom}}
    \end{minipage}
\end{figure}

\textbf{Magnetic Rotary Encoder.}
We propose a low-cost AS5600 magnetic rotary encoder solution to achieve accurate and robust gripper state capture.
As shown in Fig.~\ref{fig:as5600}, we modify the top cover of UMI to attach a radial magnet to one of the joints of the mechanical assembly, with the Hall sensor positioned above it at an appropriate distance ($\sim$ 2 mm).
The AS5600 provides high-resolution 12-bit position readings (4,096 positions per revolution) and communicates with the single-board computer through the I$^2$C protocol. This solution offers a higher sampling rate and resolution, immunity to visual occlusion, and negligible computational overhead.

\textbf{AR Motion Capture System.}
To overcome the limitations of vision-based tracking (SLAM) in 
\begin{wrapfigure}{r}{0.43\linewidth}
    \centering
    \includegraphics[width=\linewidth]{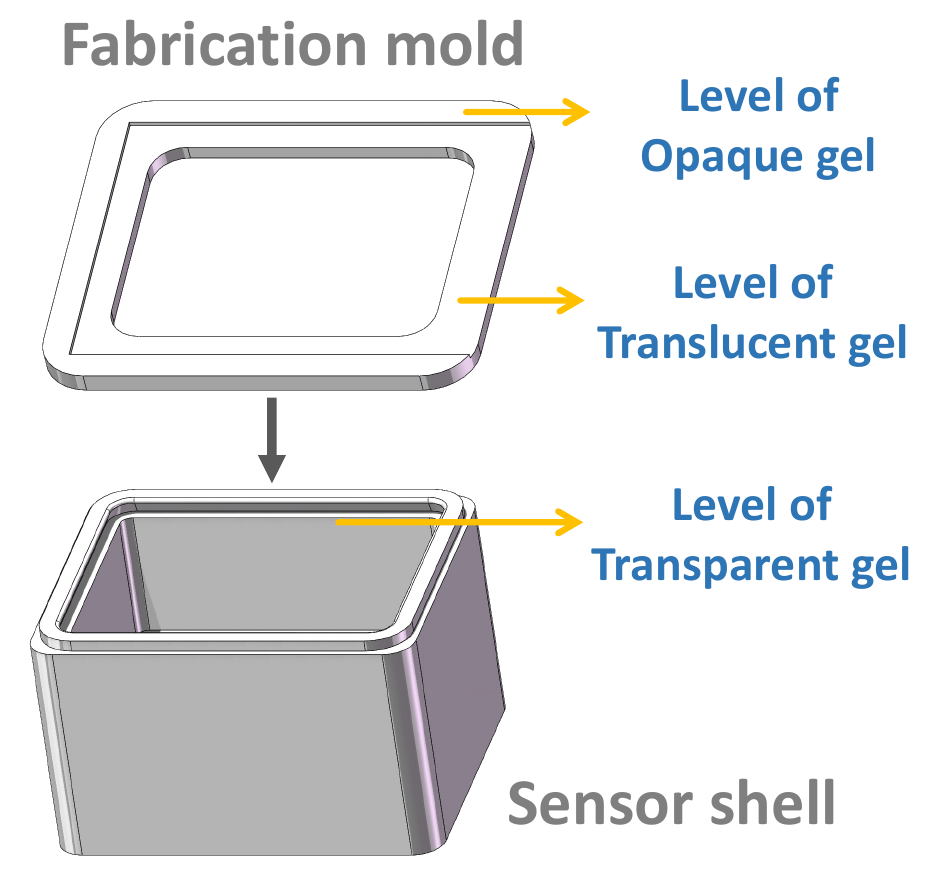}
    \caption{The mold for stable fabrication of the tactile sensor.}
    \label{fig:9dtact-mold}
\end{wrapfigure}
occluded or complex scenarios, we adopt an AR-based approach for end-effector pose estimation. 
Following ARCap~\cite{chen2024arcap}, our system uses a Meta Quest 3 headset for 6D pose motion capture, which is accurate and robust to occlusion.
We integrate the left VR controller through a custom-designed mount attached to the UMI body, and use the headset to track the 6D pose of the controller.
The mount also provides additional space for the power supply and an Orange Pi controller, serving as a universal sensor hub, synchronously capturing data from the AR headset, rotary encoder, and any additional sensors, such as tactile sensors. 
The tracking range of the system is 3 meters, sufficient for tasks at a typical robot workspace. While our system can also easily adapt to other trackers (\textit{e.g.} HTC Vive trackers), our goal of using VR is the real-time evaluation of tracking and future extension of a user-friendly interface to guide the crowdsourcing.
\\
The orientation of the VR controller is arbitrary since the transformation between the controller and UMI coordinate frames will be determined through our calibration pipeline.
This flexible mounting approach simplifies the assembly and avoids precise physical alignment.

\textbf{Visual Input.}
Same as the UMI~\cite{umi}, we employ a GoPro camera with a fisheye lens as our primary visual input. We move the camera forward, following FastUMI~\cite{wu2024fast} for a wider and clearer view. The camera positioning eliminates body occlusion in the field of view to enhance transferability.

\textbf{Fingertip Visuo-tactile Sensors.}
We improve 9D-Tact~\cite{9dtact} as a low-cost and DIY-friendly tactile senser for exUMI.
We use 9DTact for low-cost. It is not as accurate as GelSight for surface geometry, but it provides sufficiently rich information about contact forces (normal+tangent) with fast deformation recovery, which is justified by our experiments.
\\
As shown in Fig.~\ref{fig:9dtact} and Fig.~\ref{fig:9dtact-mold}, our enhancements involve:
(1) We redesigned the sensor shell to securely anchor the top silicone layer, enhancing its resistance to tangent forces and ensuring long-term stability.
(2) The LED board was modified to incorporate a more robust 2-pin header connector for stable power delivery. Compared to USB cables, Dupont connectors offer superior cable management flexibility. Plus, the LEDs were rearranged to minimize power consumption—a critical factor for our embedded system’s efficiency.
(3) A custom mold was developed to precisely control the silicone layer’s thickness. The mold is affixed to the sensor shell, allowing controlled pouring of transparent, translucent, or opaque liquid silicone until it reaches the desired level (Fig.~\ref{fig:9dtact-mold}). Excess silicone is then removed by carefully scraping along the mold’s surface with a spatula.
\\
The upgraded sensor design achieves significantly enhanced durability and stability. Please refer to the appendix for more fabrication details.

\textbf{Cost and Accessibility.}
We show the overall bill of materials in Tab.~\ref{tab:bom}. Our system is low-cost with a minimal configuration starting at \textbf{\$ 698}, which can be further reduced by substituting the GoPro with alternative fisheye cameras. The battery duration of Meta Quest headset is around 4 hours, and that of the Orange Pi system is over 10 hours.
Our design is DIY-friendly and use readily available components, making it suitable for research and education. All CAD files will be released.

\subsection{Data Collection and Processing}

\textbf{AR Capture Interface}.
Building upon the remarkable engineering of ARCap~\cite{chen2024arcap}, we simplify the 
\begin{wrapfigure}{r}{0.32\linewidth}
    \centering
    \vspace{-2mm}
    \includegraphics[width=0.8\linewidth]{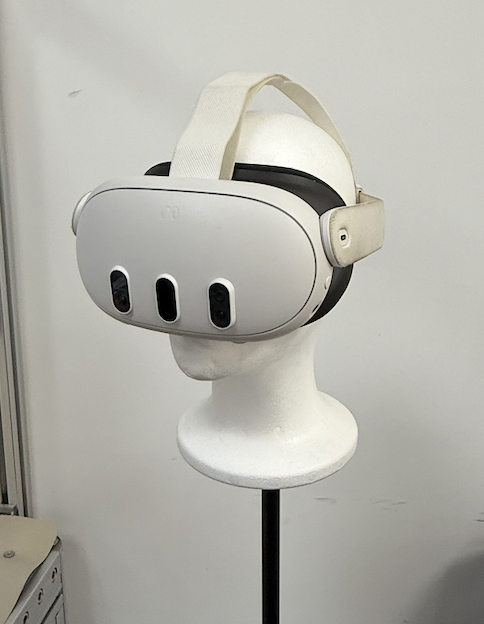}
    \caption{VR Headset Stand}
    \label{fig:vr-stand}
    \vspace{-5mm}
\end{wrapfigure}
socket-based data transfer interface for the 6D pose capture process. The collection procedure is as follows:
\begin{enumerate}[leftmargin=3em,itemsep=0pt,topsep=0pt]
\item Initialize the server program on the Raspberry Pi.
\item Launch the client application on the Meta Quest headset.
\item Set up the base coordinate frame in AR space. Then the headset can be \textbf{optionally} placed on a stand for convenience.
\item Begin data streaming of real-time 6D controller poses to the Raspberry Pi.
\end{enumerate}

\textbf{Calibration of AR Latency.}
To synchronize AR motion capture with the visual inputs, we designed a calibration protocol involving horizontal sweeps in front of a stationary ArUco marker. We extract the x-axis movement of the AR MoCap system and the camera-detected marker trajectories, then use a bisection-style optimization algorithm to align the two trajectories and compute the latency of the AR system. 
Specifically, given the two 1-dimension trajectories on timesteps $\{t_i\}_{i=1}^{T}$, we convert them to two function $f(t)$ $g(t)$ \textit{w.r.t} time $t$ by interpolation, which is then calibrated by minimizing the MSE error between the two trajectory. The details are given in Alg.~\ref{alg:align}.

\begin{algorithm}[t]
    \renewcommand{\algorithmicrequire}{\textbf{Input:}}
    \renewcommand{\algorithmicensure}{\textbf{Output:}}
    \caption{Latency alignment algorithm}
    \label{alg:align}
    \begin{algorithmic}[1]
    \REQUIRE Trajectories $f(t)$ and $g(t)$, timesteps $\{t_i\}_{i=1}^{T}$, bounds of latency $\delta_{min}$ $\delta_{max}$
    \REQUIRE Constants: $\epsilon=0.0001$, search window $N$, search splits $M$
    \ENSURE Latency $\delta^*$ of $g(t)$ such that $f(t)\approx g(t+\delta^*)$
    \REPEAT
        \STATE Interpolate the interval $[\delta_{min}, \delta_{max}]$ into M segments: $\delta_0, \delta_1, \cdots,\delta_M$
        \STATE $k=\min_k \sum_{i=1}^{T}\left\| f(t_i)-g(t_i+\delta_k)\right\|_2^2$
        \STATE $\delta^* = \delta_k$  
        \STATE $\delta_{min}, \delta_{max} \gets \delta_{k-N}, \delta_{k+N} \quad$ (update the search range to the neighborhood of $\delta_k$)
    \UNTIL{$\delta_{max}-\delta_{min}<\epsilon$}
    \end{algorithmic}
\end{algorithm}

\textbf{Data Collection and Processing Pipeline.}
Compared to the original UMI system, our data collection and processing pipeline is significantly simplified and more robust:
\begin{enumerate}[leftmargin=3em,itemsep=0pt,topsep=0pt]
\item Set up the desired environment.
\item Initialize AR tracking system on the Raspberry Pi.
\item Record latency calibration sequence (one video).
\item Record demonstration videos.
\item Calculate and apply temporal latency correction.
\item Align AR capture data to the video frames through interpolation.
\item Pack synchronized data.
\end{enumerate}

\begin{figure}[t]
    \centering
    \begin{minipage}[c]{0.32\linewidth}
        \centering
        \includegraphics[width=\linewidth]{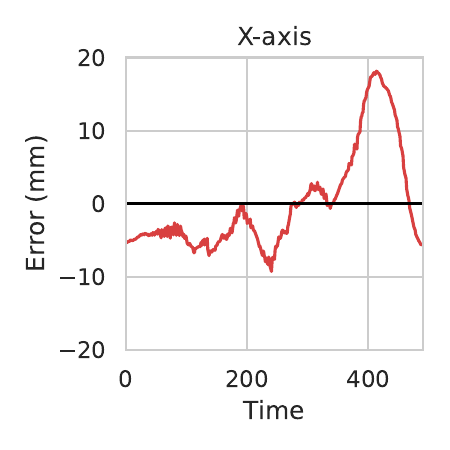}
    \end{minipage}
    \hfill
    \begin{minipage}[c]{0.32\linewidth}
        \centering
        \includegraphics[width=\linewidth]{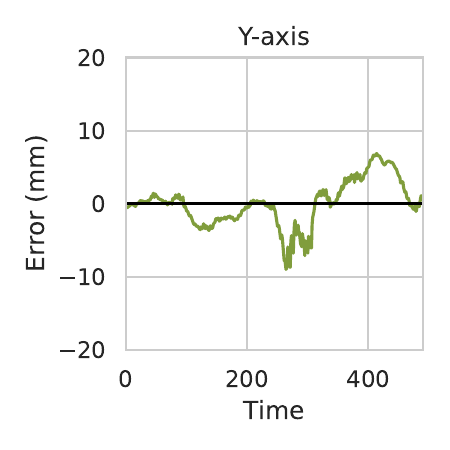}
    \end{minipage}
    \hfill
    \begin{minipage}[c]{0.32\linewidth}
        \centering
        \includegraphics[width=\linewidth]{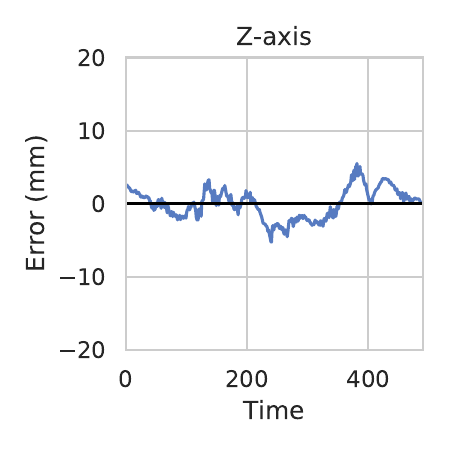}
    \end{minipage}
    \caption{Comparison of AR MoCap trajectory and ground truth trajectory.}
    \label{fig:6dpose-eval}
\end{figure}

\subsection{System Evaluation}

\textbf{Proprioception Precision}.
We first evaluated the precision of our proprioception system, particularly for the AR-based MoCap system. 
We obtained both ground truth and AR controller trajectories by mounting the AR controller on the robot end-effector and teleoperating the robot. 
We evaluate the system by moving the robot within a 50 cm range. The resulting 6D pose differences between the estimated and ground truth values are presented in Fig.~\ref{fig:6dpose-eval}
The system demonstrates remarkable accuracy, achieving mean position errors of 5.4 / 2.3 / 1.7 mm at each axis. The rotation errors are below 1 degree (notably small due to the robot's limited rotation range). The x-axis error reaches a maximum of 20 mm since it is the depth axis in the Flexiv coordinate system and inherently presents greater measurement challenges. These high-precision measurements enable efficient robot policy learning by providing high-quality training data.

\begin{figure}[t]
    \begin{minipage}[t]{0.41\linewidth}
        \centering
        \includegraphics[width=\linewidth]{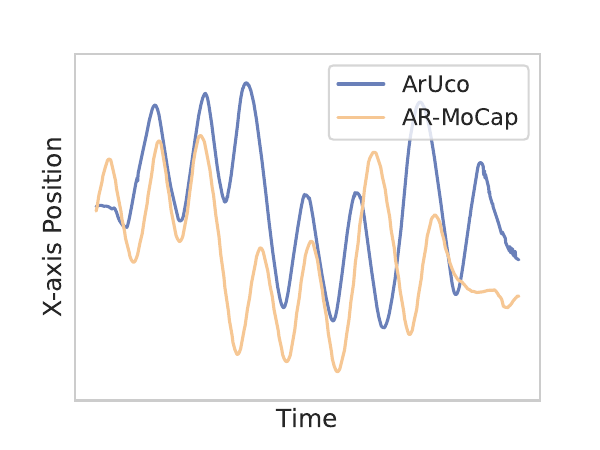}
    \end{minipage}
    \hfill
    \begin{minipage}[t]{0.41\linewidth}
        \centering
        \includegraphics[width=\linewidth]{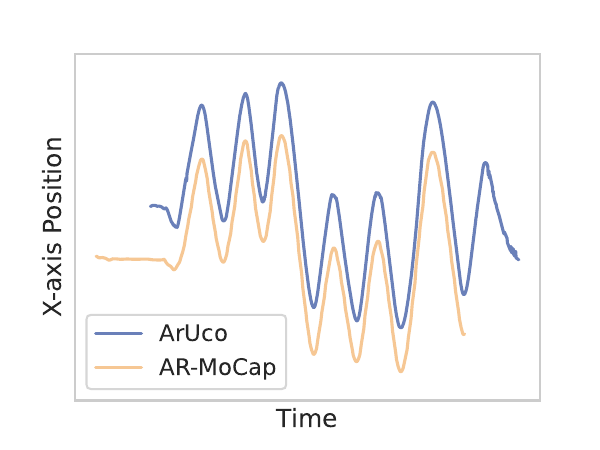}
    \end{minipage}
    \caption{An example of AR MoCap and Vision trajectory before (left) and after (right) alignment.}
    \label{fig:latency-real}
\end{figure}

\textbf{Latency Calibration}.
We give a qualitative visualization of our latency calibration algorithm in Fig.~\ref{fig:latency-real}. The x-axis trajectory of the AR MoCap system and the visual input (represented as the trajectory of the ArUco marker) are perfectly aligned after our calibration system, and we can read the time offset for further modality alignment. With proper sweeping frequency, our system could consistently achieve less than 5 ms latency error.

\begin{tcolorbox}[colback=gray!10, colframe=black, sharp corners]
\textbf{\large Key Takeaways}

We provide an accurate, extensible, cost-effective, and DIY-friendly enhancement to the UMI system. Our solution is particularly valuable if you need:

$\bullet$  \textbf{Enhanced Tracking Precision}: Achieve $\sim$100\% data effective ratio in diverse environments by our AR-based MoCap system and the magnetic rotary encoder for gripper state.\\ 
$\bullet$  \textbf{Non-Parallel Gripper Support}: An alternative design is provided for more popular industrial grippers like Flexiv Grav or Robotiq 2F series with an adaptable mechanical design.\\
$\bullet$  \textbf{Multimodal Sensing}: Including tactile, audio, or other custom sensors through our modular hardware interface.

All components are commercially available, and the fabrication details will be opened soon.
\end{tcolorbox}

\section{Detailed Taxonomy of Tactile Representation Learning}  
\label{app:sec:taxonomy}

We give more details of our discussion on the current taxonomy of tactile representation learning.

The target of tactile representation learning is to learn a tactile encoder $\mathcal{E}_T$ for the tactile data, to facilitate further multimodal policy learning:
\begin{equation}
\label{eq:tactile-policy}
    \pi(\mathbf{a}_{t+1} | \mathcal{E}_S(\mathbf{s}_{t}), \mathcal{E}_T(\mathbf{T}_t), \mathcal{E}_V(\mathbf{V}_t)).
\end{equation}
Current tactile representation methods fall into three dominant paradigms, each with specific advantages and fundamental constraints:

\textbf{(a) Direct Multimodal Imitation Learning}~\cite{10611113,Lee_2024,10802778,10610567,pattabiraman2024learningprecisecontactrichmanipulation}.
This approach trains end-to-end multimodal policies in Eq.~\ref{eq:tactile-policy} using paired tactile-visual-action data, and learn the tactile representation $\phi_T$. While effective for narrow tasks, it suffers from tactile data scarcity, since the tactile contacts only occur in very few frames in most regular robot tasks (\eg $\sim3000$ frames in 100 demos of pick and place task). The method also inherently couples task objectives with tactile features, limiting cross-task transferability. 

\textbf{(b) Spatial Self-Supervised Learning}~\cite{rodriguez2024contrastivetouchtotouchpretraining,guzey2023dexteritytouchselfsupervisedpretraining,yu2025mimictouchleveragingmultimodalhuman,wu2025canonicalrepresentationforcebasedpretraining}.
Self-supervised learning (SSL) is adopted for a generic and transferable tactile representation and learn tactile embeddings $\mathcal{E}_T(\mathbf{T}_t)$ through proxy objectives on task-agnostic unlabeled data. Mostly, SSL is adopted for tactile pretraining \textbf{spatially}, \ie treating tactile frames as images and applying image-level SSL algorithms. However, these spatial SSL methods potentially exploit incorrect inductive bias for tactile learning.
Methods like contrastive learning~\cite{rodriguez2024contrastivetouchtotouchpretraining,guzey2023dexteritytouchselfsupervisedpretraining,yu2025mimictouchleveragingmultimodalhuman} usually assumes translation invariance, which may not exist in tactile sensing: any translation or shifting of tactile frames result in different contact point information.
Similarly, masked learning~\cite{wu2025canonicalrepresentationforcebasedpretraining,higuera2024sparsh,feng2025anytouch} methods assumes that some image patches could be recovered from other patches, but tactile images do not hold geometrical self-consistency: \eg, a three-finger press and a four-finger press on a tactile sensor produce different signals, but partial masking could produce an identical image showing only two contact points. This ambiguity makes it impossible to recover the correct original input from the masked data.
Therefore, it is challenging to design a self-supervising proxy objective for a robot task.

\textbf{(c) Visual-Tactile Alignment}~\cite{VisGel,clarke2025x,feng2025anytouch}.
Cross-modal alignment learns joint embeddings by maximizing in-pair similarity $s\left(\mathcal{E}_T(\mathbf{T}_t), \mathcal{E}_V(\mathbf{V}_t)\right)$  of visual and tactile modalities. Though it is effective for visual-language learning, it fundamentally assumes a coarse \textit{one-to-one visuo-tactile mapping}, regardless of the actual \textit{one-to-many relation}: with  different contact forces, identical visual scenes yield divergent tactile signals. The multimodal alignment also overlooks that visual and tactile sensing are complementary rather than well-aligned. For robot learning, tactile sensors are a complementary information to the visual input, but the alignment method discards this privileged information.

Beside these approaches, some methods use pure generative models to learn a compact latent that preserves most tactile clues, such as auto-encoder~\cite{zhao2024transferable} or VQ-GAN~\cite{xu2025unit}, which is more reasonable.
To take a step further, we consider the proxy task of temporal prediction, by reformulating tactile representation as an action-conditioned temporal prediction problem, explicitly modeling the forward tactile dynamics that underpin real-world contact interactions.

\begin{figure}[t]
    \centering
    \begin{minipage}[b]{0.7\linewidth}
        \centering
        \resizebox{1.0\linewidth}{!}{
        \begin{tabular}{lcccc}
        \toprule
        Dataset     & \textbf{Data Scale} & \textbf{Tactile Sensor} & \makecell{\bf Proprioception \\ \bf / Action}  & \makecell{ \bf Collection \\ \bf Source}  \\
        \midrule
        \citet{MoreThanFeel}                   &  6.5 K &  GelSight & \cmark &  Robot    \\
        \citet{FeelSucc}                       &  9.3 K &  GelSight & \cmark &  Robot    \\
        VisGel~\citep{VisGel}                  & 12.0 K &  GelSight & \cmark &  Robot    \\
        \citet{burka2018inst}                  &  1.1 K &  Multiple & \cmark &  Human    \\ 
        Touch and Go~\citep{yang2022touchgolearninghumancollected}     & 13.9 K &  GelSight & \xmark &  Human    \\
        ObjectFolder Real~\citep{objfolder-r}  &  3.0 K &  GelSight & \cmark &  Robot    \\
        SSVTP~\citep{kerr2022self}             &  4.5 K &   DIGIT   & \cmark &  Robot    \\
        TVL~\citep{fu2024touch}                & 43.7 K &   DIGIT   & \cmark &  Robot    \\ 
        Touch2Touch~\citep{touch2touch}        & 32.3 K &  Multiple & \cmark  & Robot  \\  
        X-Capture~\citep{clarke2025x}          &  3.0 K &   DIGIT   & \xmark &  Human    \\
        \midrule
        \textbf{Ours}                         & \makecell{480.9 K \\ (raw frames)} &  9DTact+  & \cmark &  Human \\
        \bottomrule
        \end{tabular}
        }
        \captionof{table}{Comparison of real-world tactile datasets.} 
        \label{tab:dataset-compare}
    \end{minipage}
    \hfill
    \begin{minipage}[b]{0.25\linewidth}
        \centering
        \includegraphics[width=\linewidth]{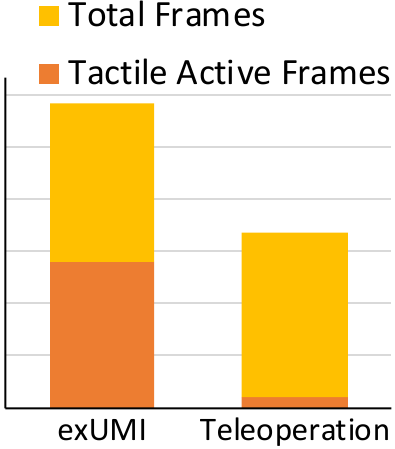}
        \caption{Comparison of per-hour tactile data collection efficiency.}
        \label{fig:data-efficiency}
    \end{minipage}
\end{figure}

\section{More Implementation Details}
\label{app:sec:details}

\textbf{Tactile Data Curation.}
Since active tactile signals are sparse in real-world data collection, we adopt a data rejection strategy during data sampling to avoid trivial samples. For each data chunk, we check the proportion of active pixels for each tactile frame. If the active proportion of all frames is below a certain threshold, the data chunk will be discarded and resampled.

\textbf{Implementation Details.}
For each timestep, our exUMI collects two tactile images on the two sides of the gripper. We convert the images to a calibrated grayscale image following 9DTact~\cite{9dtact}, and extract the convex and concave pixel map by comparing the grayscale image to the reference image (tactile signal at no contact). The grayscale image, convex map, and concave map are stacked as a 3-channel image for a richer representation of tactile contacts. 

We pretrain the tactile representation on our large-scale human play dataset, which is randomly split into a training and validation set by 15:1.
The action sequence is represented as the relative pose and the gripper state. The images on two sides are concatenated and then downsampled to 224$\times$224 resolution.
We use a pretrained VAE model (KL-F16) as the encoder and decoder for tactile learning.
The tactile prediction is conducted in 8 temporal frames, where 4 random frames in the first half are regarded as input and 4 frames in the second half are the prediction target. We adopt a larger frequency for action following \citet{uva}.
The tactile prediction model reaches quick convergence due to the simpler distribution of tactile sensing.

\begin{figure} 
    \centering
    \includegraphics[width=0.5\linewidth]{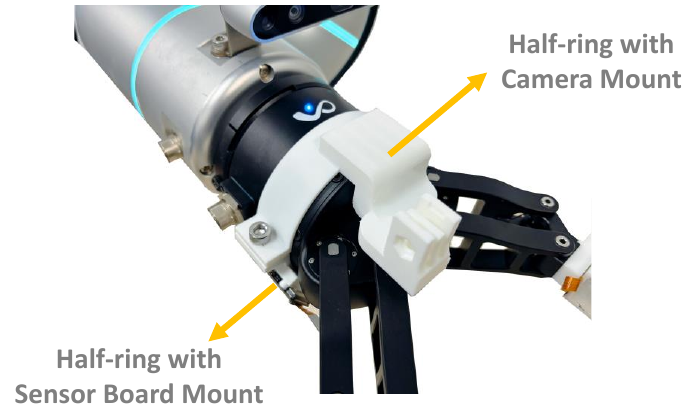}
    \caption{Pipe clamp style GoPro mount for deployment.}
    \label{fig:deploy}
\end{figure}

\textbf{Environment Details}.
The camera and sensors are connected to a master computer with a RTX 4070 GPU, which controls the robot at a frequency of 10 Hz.
As illustrated in Fig.~\ref{fig:deploy}, we designed a pipe clamp-style camera mount and gripper tactile sensor mount to exactly replicate the end-effector sensor placement of the exUMI system. The mount consists of two half-rings, one integrated with a standard GoPro mount. 
This design can be attached to the Flexiv Grav gripper base, and also adaptable on various other end effectors.

\textbf{Task Details}.
(1) {\it Pick cube / carrot / broccoli}: picking up an object and place it into a container. The object and the target container are randomized within a 30cm$\times$30cm area.
(2) {\it Insert pen}: moving a pen from one cup to another cup. The cups are randomized within a 30cm$\times$30cm area, and the pen is randomly placed in the cup. The colors are fixed.
(3) {\it Stack cubes}: stacking a 5cm$\times$5cm$\times$5cm cube on top of another, requiring precise manipulation ability. The cubes are randomized within a 30cm$\times$30cm area. The colors are fixed.

For tactile-aware policy learning, we evaluate on more complex tasks:
(1) {\it Put Ball}: pick up a yellow soft ball (radius$=3$cm) and place it in a red cup. The cubes are randomized within a 30cm$\times$30cm area.
(2) {\it Open Bottle}: rotate the bottle cap until it is fully unscrewed. The bottle has diameter$=6.5$cm and height$=13$cm, and is randomized within a 20cm$\times$20cm area.
(3) {\it Pull Drawer}: pull out a drawer, which is either empty (``Empty'') or contains a random amount of stones (``Random'', randomized from 50g to 1000g), requiring tactile clues to determine the pulling direction.
(4) {\it Peg in hole}: insert a 4 cm $\times$ 3 cm yellow block into a 4.3 cm $\times$ 3.3 cm slot, requiring a precision and force-aware adjustment. The yellow block is randomly put on a 10cm$\times$10cm blue cube. We split the task into ``Grasp'' and ``Insert'' stages.

\begin{figure}[t]
    \centering
    \begin{minipage}[c]{0.55\linewidth}
        \centering
        \resizebox{\linewidth}{!}{
            \begin{tabular}{l|cccc}
            \toprule
            Modality        &  V   &   V+T  &  V+T & V+T      \\
            Tactile Learning&  /   & Direct & BYOL & TPP (Ours) \\
            \midrule    
            Put Ball        & 70\% &  70\%  & 80\% & \bf 85\%      \\
            Peg in Hole     & 50\% &  60\%  & 50\% & \bf 80\%    \\
            \bottomrule
            \end{tabular}
        }
        \captionof{table}{Real world evaluation of tactile represention learning algorithms. V: vision; T: tactile.}
        \label{tab:tactile-learning-more}
    \end{minipage}
    \hfill
    \begin{minipage}[c]{0.4\linewidth}
        \centering
        \includegraphics[width=\linewidth]{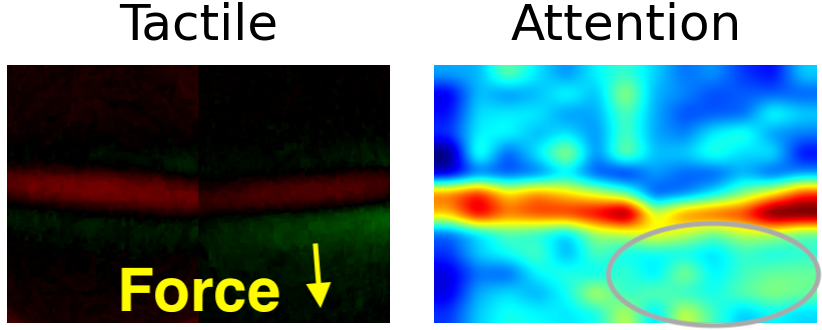}
        \caption{Attention Map.}
        \label{fig:attention}
    \end{minipage}
\end{figure}

\section{Ablation Experiments}

\subsection{Attention Visualization}

We visualize the attention map of tactile encoder pretrained by TPP. We give an example in Fig.~\ref{fig:attention} on ``pull drawer'' task, where the arrow shows the tangent force direction.
The pretrained tactile model focuses on the area that indicates force magnitude (red area), and also the direction (in the circle).

\subsection{Tactile Learning Comparison}

To compare our representaion learning algorithm, we implement the direct learning method, and a spatial self-supervised learning method BYOL following MimicTouch~\cite{yu2023mimictouch}, which is pretrained on our collected tactile dataset. The results are shown in Tab.~\ref{tab:tactile-learning-more}. TPP shows the best success rate on two tasks.

\section{Broader Impact}

The open-source design and affordability (\$698, and we are working on making it less than \$500) of the exUMI system democratize tactile robotics research by lowering technical and financial barriers for resource-constrained labs and educational institutions. This accessibility can accelerate innovation and broaden participation in the field. Beyond the research community, this work has direct societal applications. In assistive robotics, it enables robots to perform delicate tasks for the elderly or individuals with disabilities. In industrial safety, sophisticated tactile sensing allows robots to operate more safely alongside human workers and enabling the reliable handling of fragile components. By making advanced tactile learning more accessible, our work helps pave the way for robots that can interact with the physical world more safely and intelligently.

\end{appendix}

\end{document}